\RequirePackage{amsthm}
 
\documentclass[pdflatex,sn-nature]{sn-jnl}

\usepackage{graphicx}%
\usepackage{multirow}%
\usepackage{amsmath,amssymb,amsfonts}%
\usepackage{amsthm}%
\usepackage{mathrsfs}%
\usepackage[title]{appendix}%
\usepackage{xcolor}%
\usepackage{textcomp}%
\usepackage{manyfoot}%
\usepackage{booktabs}%
\usepackage{algorithm}%
\usepackage{algorithmicx}%
\usepackage{algpseudocode}%
\usepackage{listings}%
\usepackage{subfigure}%
\usepackage{cleveref}

\usepackage{lmodern}%
\usepackage[super]{nth}%

\newcommand{\R}{\mathbb{R}}

\theoremstyle{thmstyleone}%
\theoremstyle{thmstyletwo}%

\theoremstyle{thmstylethree}%
\newtheorem{definition}{Definition}%

\graphicspath{{figures/}}

\begin{document}

\title[The impact of behavioral diversity in multi-agent reinforcement learning]{The impact of behavioral diversity in multi-agent reinforcement learning}



\author*[1]{\fnm{Matteo} \sur{Bettini}}\email{mb2389@cl.cam.ac.uk}

\author[1]{\fnm{Ryan} \sur{Kortvelesy}}\email{rk627@cl.cam.ac.uk}

\author*[1]{\fnm{Amanda} \sur{Prorok}}\email{asp45@cl.cam.ac.uk}

\affil[1]{\orgdiv{Department of Computer Science and Technology}, \orgname{University of Cambridge}} 

\abstract{ 
Many of the world’s most pressing issues, such as climate change and global peace, require complex collective problem-solving skills.
Recent studies indicate that diversity in individuals’ behaviors is key to developing such skills and increasing collective performance.
Yet behavioral diversity in collective artificial learning is understudied, with today’s machine learning paradigms commonly favoring homogeneous agent strategies over heterogeneous ones, mainly due to computational considerations.
In this work, we employ diversity measurement and control paradigms to study the impact of behavioral heterogeneity in several facets of multi-agent reinforcement learning.
Through experiments in team play and other cooperative tasks, we show the emergence of unbiased behavioral roles that improve team outcomes; how behavioral diversity synergizes with morphological diversity; how diverse agents are more effective at finding cooperative solutions in sparse reward settings; and how behaviorally heterogeneous teams learn and retain latent skills to overcome repeated disruptions.
Overall, our results indicate that, by controlling diversity, we can obtain non-trivial benefits over homogeneous training paradigms, demonstrating that diversity is a fundamental component of collective artificial learning, an insight thus far overlooked.
}





\maketitle

\section{Introduction}\label{sec:intro}

Diversity of skills and behaviors is ubiquitous in real life and is widely believed to be key to the survival and thriving of natural ecosystems~\cite{cardinale2012biodiversity}.
It enables collective intelligence~\cite{kameda2022information, woolley2015collective}, a property that is not simply dependent on the maximum or total intelligence of team members.
Collective intelligence in nature emerges from individuals learning in rich and heterogeneous environments, surrounded by diverse functional stimuli and interactions~\cite{baumann2024network}.
Increasing evidence, however, suggests that we may have underestimated the impact of diversity in the learning of behaviors that solve complex cooperative tasks~\cite{chenghao2021celebrating,wang2020roma}.
In spite of this, current research in collective artificial learning has left diversity largely underexplored.

This issue is particularly evident in the context of Multi-Agent Reinforcement Learning (MARL), which is typically applied to synthesize team strategies in such collective learning problems. 
Traditional MARL algorithms constrain the strategies (i.e., policies) of agents to be identical~\cite{gupta2017cooperative,rashid2018qmix,sukhbaatar2016learning}.
This speeds-up learning by training a shared policy from all individuals' experiences, but results in the agents becoming behaviorally homogeneous.
While some MARL methods encourage behavioral diversity~\cite{jiang2021emergence, chenghao2021celebrating,mahajan2019maven,wang2020roma}, they blindly promote it via additional learning objectives, lacking principled techniques to measure and control it. 
Despite these efforts, the benefit of diversity towards the resolution of complex cooperative tasks remains poorly understood. 
To address this, in prior work we introduced principled techniques that enable the measurement and exact control of behavioral diversity in agent teams~\cite{bettini2023snd,bettini2024controlling}, enabling new avenues for the study of heterogeneity.

In this work, we employ these diversity measurement and control paradigms to study how and in what form behavioral diversity contributes to collective artificial learning.
We decompose collective learning into three core challenges: (1) team play, the ability to cooperate with teammates in long-term decision-making; (2) exploration, the ability to discover the world and task objectives; (3) resilience, the ability to recover from unexpected disruptions.
Our study spans several cooperative tasks including soccer, to evaluate team play, collective foraging, to evaluate exploration, and physically-coupled multi-agent navigation in the presence of dynamic disruptions, to evaluate resilience.
We show that diversity is responsible for the emergence of complementary roles and better performing strategies in team play, that it boosts the learning process through more efficient exploration, and finally, that it enables agents to learn and retain latent skills to overcome repeated disruptions.
Via this study, we aim to demonstrate that behavioral diversity should not be overlooked due to computational overheads, but should rather be considered a key enabler of collective learning.

\begin{figure*}[!tp]
\centering
\includegraphics[width=\textwidth]{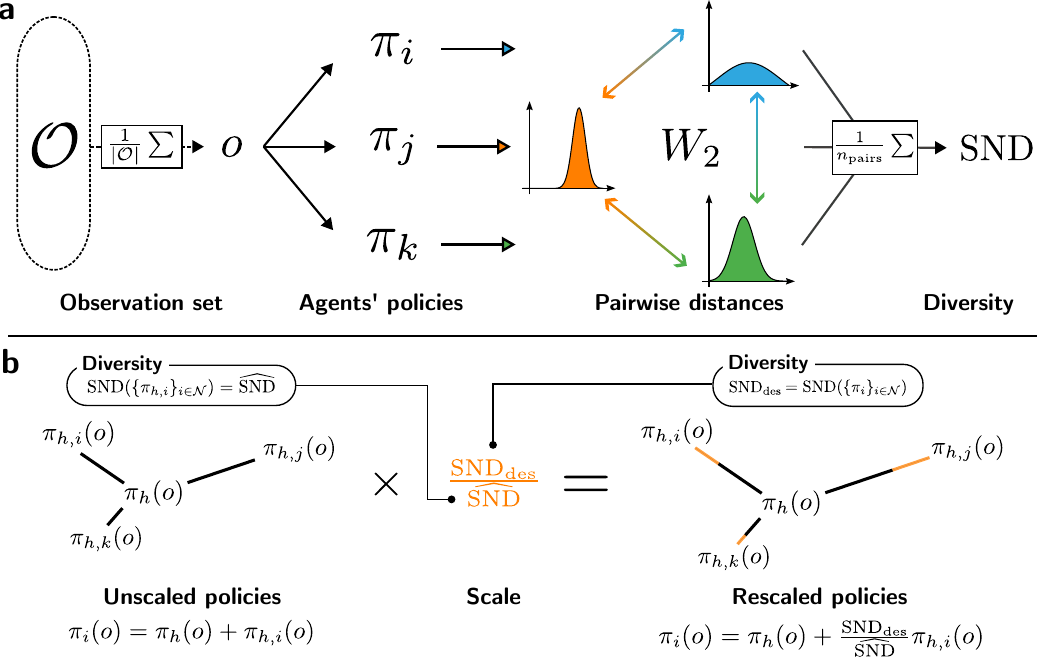}
\caption{\textbf{Overview on measuring and controlling diversity.}
\textbf{a,} Diversity ($\mathrm{SND}$) is computed as the average Wasserstein distance between agents' action distributions over all agent pairs and all observations in the evaluation set.
\textbf{b,} Diversity is controlled by
representing policies as the sum of a shared (homogeneous) component and individual (heterogeneous) components, which are dynamically scaled according the the
current ($\widehat{\mathrm{SND}}$) and desired ($\mathrm{SND}_\mathrm{des}$) value of the diversity metric.
}
\label{fig:intro}
\end{figure*}

\subsection{Measuring diversity}
We employ System Neural Diversity ($\mathrm{SND}$)~\cite{bettini2023snd} to measure heterogeneity.
Given a team of agents $\mathcal{N}=\{1,\ldots, i,\ldots, n\}$, agent policies are functions $\pi_{i}(o_i)$ that produce a continuous action distribution given an observation $o_i$.
We measure the teams' diversity in two phases. First, we employ the Wasserstein statistical metric~\cite{vaserstein1969markov} to measure the diversity among agent pairs over a set $O$ of observations: $d(\pi_{i},\pi_{j}) = \frac{1}{|O|}\sum_{o\in O }W_2(\pi_{i}(o),\pi_{j}(o))$. Pairwise behavioral distances are then aggregated in a system-level metric by taking the mean over agent pairs:
\begin{equation}
    \mathrm{SND}:= \mathrm{SND}(\left \{ \pi_{i} \right \}_{i \in \mathcal{N}}) = \frac{2}{n(n-1)|O|}\sum_{i=1}^n\sum_{j=i+1}^n \sum_{o\in O }W_2(\pi_{i}(o),\pi_{j}(o)).
    \label{eq:snd}
\end{equation}
This process (depicted in \autoref{fig:intro}a) quantifies behavioral diversity as the behavioral dispersion of the system.

\subsection{Controlling diversity}
To control diversity, we employ Diversity Control (DiCo)~\cite{bettini2024controlling}, which allows to constrain the diversity of a multi-agent system to a desired metric value $\mathrm{SND}_\mathrm{des}$.
This method works by representing each agent's policy $\pi_i(o)$ as the sum of a \textit{homogeneous parameter-shared component} $\pi_{h}(o)$ and a \textit{per-agent heterogeneous deviation} $\pi_{h,i}(o)$.
The per-agents policy deviations are then rescaled with respect to the homogeneous component to obtain the desired diversity. The scaling factor is computed by dividing by their current diversity ($\widehat{\mathrm{SND}} := \mathrm{SND}(\left \{ \pi_{h,i} \right \}_{i \in \mathcal{N}})$) and multiplying by the desired one ($\mathrm{SND}_\mathrm{des}$):
\begin{equation}
     \pi_i(o) = \pi_h(o) + \frac{\mathrm{SND}_\mathrm{des}}{\widehat{\mathrm{SND}}}\pi_{h,i}(o).
    \label{eq:dico}
\end{equation}
This process (depicted in \autoref{fig:intro}b) guarantees that the diversity of the multi-agent system will be equal to the desired one $ \mathrm{SND}(\{\pi_{i}\}_{i\in\mathcal{N}}) = \mathrm{SND}_\mathrm{des}$.

\section{Results}\label{sec:results}

\subsection{Behavioral diversity is key to team play}
\begin{figure*}[!pt]
\centering
\includegraphics[width=\textwidth]{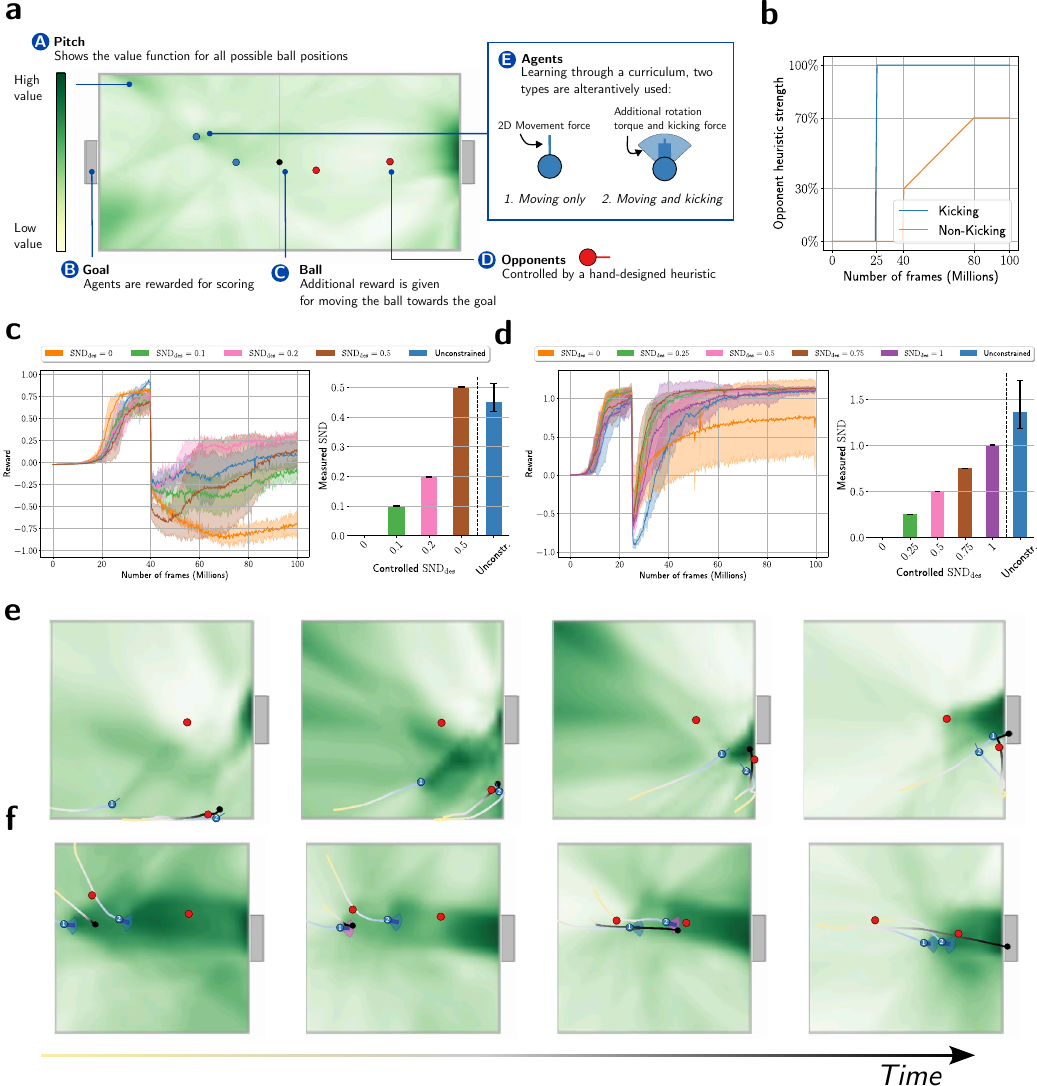}
\caption{
\textbf{Soccer results in the two vs. two setting.} 
\textbf{a,} Setup and details of the \textit{Soccer} scenario.
\textbf{b,} Curriculum used for opponents' strength annealing in the two experiment setups considered.  
\textbf{c,d,} Reward (throughout training, with $1$ obtained when agents score in $100\%$ of the matches) and diversity ($\mathrm{SND}$, after training) for \textit{non-kicking} (\textbf{c}) and \textit{kicking} (\textbf{d}) agents. We report results for agents constrained at different diversity levels (including homogeneous, $\mathrm{SND}_\mathrm{des}=0$) and unconstrained heterogeneous agents. 
The results show that agents constrained to a high diversity obtain the best performance, while unconstrained agents tend to become too diverse, not leveraging any homogeneity. 
\textbf{e,f,} Renderings from \textit{non-kicking} (\textbf{e}) and \textit{kicking} (\textbf{f}) heterogeneous agents, showing the emergence of diverse strategies that resemble a \textit{crossing pass} (\textbf{e}) and a \textit{through pass}  (\textbf{f}).
Homogeneous agents are not able to learn such strategies and often converge to collectively blindly chase after the ball (a known suboptimal policy).
All experiments are run for 3 random seeds.
Each datapoint in the reward curves is computed over 480 matches ($\times 3$ seeds). Each match lasts max 500 steps or until a team scores. Reward curves report mean and standard deviation, while bar charts report mean with error bars representing the \nth{25} and \nth{75} percentiles.
}\label{fig:2v2}
\end{figure*}

Team play, or cooperation, requires the collective ability to tackle a shared objective with mutual assistance of team members. As such, it provides an important benchmark for collective intelligence in both human and artificial systems, and has served as a milestone demonstrator for methodological progress~\cite{sebanz2006joint,ijspeert2001collaboration,deepminssimulationsoccer}. 
To evaluate team play in collective learning, we train agents with a \textit{shared reward}.\footnote{The diversity in this setting corresponds to $\mathcal{B}_s$-heterogeneity, according to the taxonomy from~\cite{bettini2023hetgppo}}. 
We begin by studying the impact of behavioral diversity when agents (i.e., team players) are \textit{physically-identical} (\autoref{sec:2v2}, \autoref{sec:5v5}, \autoref{sec:final_match}), for which the benefits of behavioral diversity have thus far been underexplored; we then focus on \textit{physically-different} agents in \autoref{sec:phy_diff}.


\subsubsection{Soccer}

We consider the game of \textit{Soccer} as it requires both high-level and low-level decision making in a long-term collective task. Despite its game-like nature, \textit{Soccer} represents one of the hardest team problems tackled by  multi-agent learning researchers to date~\cite{deepminssimulationsoccer, bipedalsoccer}. Further, due to its approachable nature with communicable results, it is an apt choice for our study.

Several existing works focus on soccer-inspired tasks.
Winners of the RoboCup soccer challenge~\cite{robocup} leverage hierarchical or hybrid learning solutions~\cite{10.7551/mitpress/4151.001.0001,MACALPINE201821} and current MARL research typically considers either small teams or simplified scenarios, all using homogeneous agents~\cite{deepminssimulationsoccer,smit2023scaling,lin2023tizero}. Our experiments are thus a departure from common practice, and represent \textbf{the first study of diversity in as large as five-player teams using end-to-end MARL}.

The setup of our \textit{Soccer} task is shown in \autoref{fig:2v2}a.
Learning agents (blue) are trained against opponents (red) executing a heuristic that is designed and tuned by hand~\footnote{Training against a heuristic opponent 
is sufficient for the goal of our study, which is to show the benefits of diversity over homogeneity under the same training regime (a fair comparison can only be guaranteed if the two paradigms are learning against the same opponent).}. Agents receive sparse rewards for scoring a goal (positive) and for conceding a goal (negative). They also receive a smaller reward for moving the ball closer to the opponent's goal.
We consider two setups: (1) \textit{non-kicking}: where agents can only move holonomically with 2D continuous action forces and have to physically touch the ball to dribble, and (2) \textit{kicking}: where agents have two additional actions for rotating and kicking the ball with a continuous force (given that the ball is within feasible range and distance). 
Agents have full observability over teammates, opponents, and the ball. 
The match terminates either when a goal is scored or when 500 timesteps have passed.

The opponents' heuristic is constructed with a high-level planner and a low-level controller that generates spline trajectories with actions of type (1) to track high-level decisions. 
The high-level policy is split into a dribbling policy (when an agent has possession) and off-the-ball movement (when it does not). 
Opponents have tunable strength parameters for: speed, decision-making (affecting off-the-ball movement and agent-ball assignment), and precision (accuracy with which it executes the plan).  
Strength parameters are updated jointly during training to create a learning curriculum, shown in \autoref{fig:2v2}b.
In particular, trained agents start by learning to score without opposition and then face the addition of opponents with the respective strength annealing.

\subsubsection{Two vs. two: diversity is fundamental to success}
\label{sec:2v2}

We begin by studying the simplest multi-agent soccer setting, consisting of a team of two players facing two opponents. 
We report reward (success rate) and learned diversity values for teams with various diversity targets ($\mathrm{SND}_\mathrm{des}$), including homogeneous teams, where $\mathrm{SND}_\mathrm{des}=0$, as well as teams for which we there is no upper bound on diversity (i.e., it is left unconstrained), for both \textit{non-kicking} (\autoref{fig:2v2}c) and \textit{kicking} (\autoref{fig:2v2}d) agents. 
In the reward plot, a score of $1$ means that learning agents score in $100\%$ of the matches, while a score of $-1$ means that the opponents score in $100\%$ of the matches (with $0$ signifying a draw on average). The rewards and the opponent strength in the \textit{kicking} setting are generally higher as the agents in this case have the ability to kick as an advantage over opponents.

At all diversity levels, both paradigms are able to learn to score without opponents in the first phase of learning. 
After the opponents are introduced, homogeneous teams exhibit lower performance compared to heterogeneous ones. This is due to the fact that they converge to locally-optimal policies; the most common resulting homogeneous strategy consists of all agents chasing after the ball (a known suboptimal strategy in soccer). 
Unconstrained diverse teams perform better, but converge to high effective diversity levels and do not attain the best performance.
Team with controlled diversity, on the other hand, find the optimal trade-off between homogeneity (that can be used for sharing low-level skills such as dribbling) and heterogeneity (that allows diversity in higher-level strategies), as shown in \autoref{fig:2v2}ef and described below.
For \textit{non-kicking} agents (\autoref{fig:2v2}c), we see how too little ($\mathrm{SND}_\mathrm{des}=0.1$) or excessive ($\mathrm{SND}_\mathrm{des}=0.5$) controlled diversity can be detrimental, but the right level ($\mathrm{SND}_\mathrm{des}=0.2$) can lead to superior strategies against a challenging opponent.
For \textit{kicking} agents (\autoref{fig:2v2}d), where diversity is higher due to the larger action space, we observe how the right level of diversity (in this case, $\mathrm{SND}_\mathrm{des}=0.75$) leads to the quickest recovery from the addition of strong opponents, demonstrating that \textbf{diversity is key to overcoming unforeseen challenges}.

\textbf{Diverse agents learn to pass.}
Upon inspecting the cause of the higher performance by diverse teams, we observe that their policies resemble known human strategies.
In \autoref{fig:2v2}ef, we report sample renderings with \textit{non-kicking} (e) and \textit{kicking} (f) agents.
\autoref{fig:2v2}e resembles a \textit{crossing pass}, where agent $1$ heads directly to the opponents' goal, disinterested in the ball, while agent $2$ dribbles past an opponent in the corner, eventually getting the ball across, where agent $1$ receives it and scores.
\autoref{fig:2v2}f resembles a \textit{through pass}, where agent $1$ kicks the ball in the estimated future position of agent $2$, that, by the time the ball arrives, has reached it and kicks the ball in the goal, avoiding the incoming opponent.
\textbf{These strategies emerge thanks to the diversity control, which forces agents to not all do the same thing concurrently (e.g., chase after the ball simultaneously), thus incentivizing more sophisticated team-plays and long-term strategies.} 

\begin{figure*}[!pt]
\centering
\includegraphics[width=\textwidth]{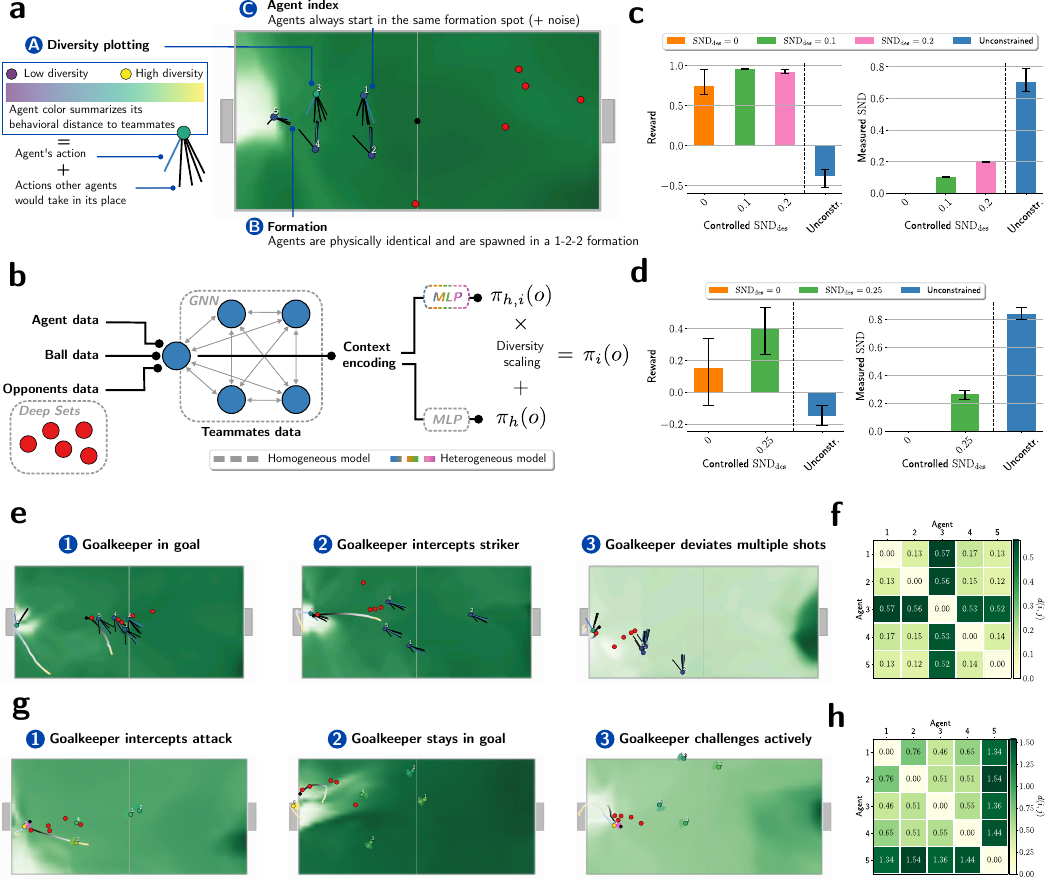}
\caption{
\textbf{Soccer results in the five vs. five setting.} 
\textbf{a,} Task setup.
\textbf{b,} Policy model used in the experiments. Deep Sets is used to grant permutation invariance over opponents' data while a Graph Neural Network (GNN) enables permutation equivariance and communication among agents. These models homogeneously compute an agent-specific context which is the input of the Diversity Control (DiCo) paradigm.
\textbf{c,d,} Training results reporting reward and diversity ($\mathrm{SND}$) for \textit{non-kicking} (\textbf{c}) and \textit{kicking} (\textbf{d}) agents.
We report results for agents constrained at different diversity levels (including homogeneous, $\mathrm{SND}_\mathrm{des}=0$) and unconstrained heterogeneous agents. 
The results show that controlling diversity grants performance improvements over homogeneous and unconstrained heterogeneous paradigms, with the latter diverging to extreme diversity and failing to solve the task. 
\textbf{e,g,} Renderings from \textit{non-kicking} (\textbf{e}) and \textit{kicking} (\textbf{g}) heterogeneous agents, showing the emergence of the \textit{goalkeeper} role.
The emergent goalkeepers exhibit preference to position themselves inside the goal independently of the context (a strategy that homogeneous agents are not able to learn) and perform multiple saves when opponents break the defense line. 
The emergence of this role is quantitatively shown in \textbf{f,h} respectively, that report pairwise agent behavioral distance $d(i,j)$ after training, showing that the goalkeepers are significantly different from all other agents. 
The reward and $\mathrm{SND}$ values are computed over 480 matches ($\times 3$ training seeds) using models trained for 150 million frames. Each match lasts max 500 steps or until a team scores. Bar charts report mean with error bars representing the \nth{25} and \nth{75} percentiles.}
\label{fig:5v5}
\end{figure*}

\subsubsection{Five vs. five: the emergence of behavioral roles}
\label{sec:5v5}

We extend the \textit{Soccer} task to a five vs. five game, with the aim to study: (1) whether the insights gathered in the 2v2 setting still hold, and (2) whether the increased number of players exacerbates the impact of diversity in teams.

We introduce further diversity visualizations (\autoref{fig:5v5}a). In particular, the color of an agent now represents its average behavioral distance from its teammates, evaluated from the point of view of that agent's observation (i.e., by how much does its action differ to the other agents' actions, were they to be in its spot). Furthermore, for \textit{non-kicking} agents, we plot the agent's action and the actions that its teammates would have taken in its stead. 

The training pipeline and curriculum remain the same as in the 2v2 setting, while the agents' policy network is changed to incorporate scalability features.
This model, shown in \autoref{fig:5v5}b, is composed of two main layers.
In the first layer, a Deep Sets~\cite{deepsets} architecture is used to process opponents' data in a permutation-invariant manner. This data, alongside ball and agent information, is then used as the node feature in a Graph Neural Network (GNN) that enables agents to communicate. The result of communication is an agent-specific context encoding. 
In the second layer, the context is given as input to the Diversity Control (DiCo) model.

Comparing the training results (\autoref{fig:5v5}cd) with the ones from the 2v2 (\autoref{fig:2v2}cd), we can confirm that the diversity benefits observed in the 2v2 extend to 5v5. 
In the \textit{non-kicking} experiments, controlled diversity in teams obtains an average score of $\approx1$, as opposed to homogeneous teams, which obtain an average score of $0.75$. 
In the \textit{kicking} experiments, now harder due to the increased player density, diverse teams still achieve the best performance.
Unconstrained diversity, however, suffers from low sample-efficiency, diverges to high effective diversity values, and is thus not able to leverage the sharing of commonly useful skills and fails to solve the task. \textbf{Although, mechanistically, the difference between controlled vs. uncontrolled diversity during learning seems subtle, they are in fact completely separate paradigms, with substantially different impact on team performance.}

\textbf{The emergence of a goalkeeper.}
The diversity-inspection tools presented in \autoref{fig:5v5}a allow us to analyze the learned strategies, and to uncover the emergence of a \textit{goalkeeper} role.
In \autoref{fig:5v5}eg, we report renderings where this role has emerged from \textit{non-kicking} (e) and \textit{kicking} (g) agents.
The pairwise agent behavioral distances (\autoref{fig:5v5}fh) for these strategies confirm that one agent becomes significantly different from the rest.
Inspecting \autoref{fig:5v5}e, where agent $3$ learns the emergent role, we observe that it learns to stay in its own goal area and wait for incoming opponents, intercepting all their attacks. The plotted actions show a strong bias to navigate to the goal, and in \autoref{fig:5v5}e3 we notice the agent's preferred left action when evaluated in all other agents' positions. The lighter agent's color further indicates its higher behavioral distance from the team.
Similar behavior can be seen in the \textit{kicking} case (\autoref{fig:5v5}g), where the emergent goalkeeper (agent $5$) also performs multiple saves, glowing of a bright yellow color, highlighting its unique role. 
The emergence of such a role is not possible for homogeneous teams where all agents are forced to take the same action given the same context.
On the other hand, diverse action strategies are commonplace for heterogeneous agents in our experiments.
\textbf{This confirms that diverse behavioral roles in soccer emerge and are beneficial even in the case of physically-identical agents. This insight would otherwise be hard to obtain from human teams, due to intrinsic (and uncontrollable) physical heterogeneity.}

\begin{figure*}[!t]
\centering
\includegraphics[width=\textwidth]{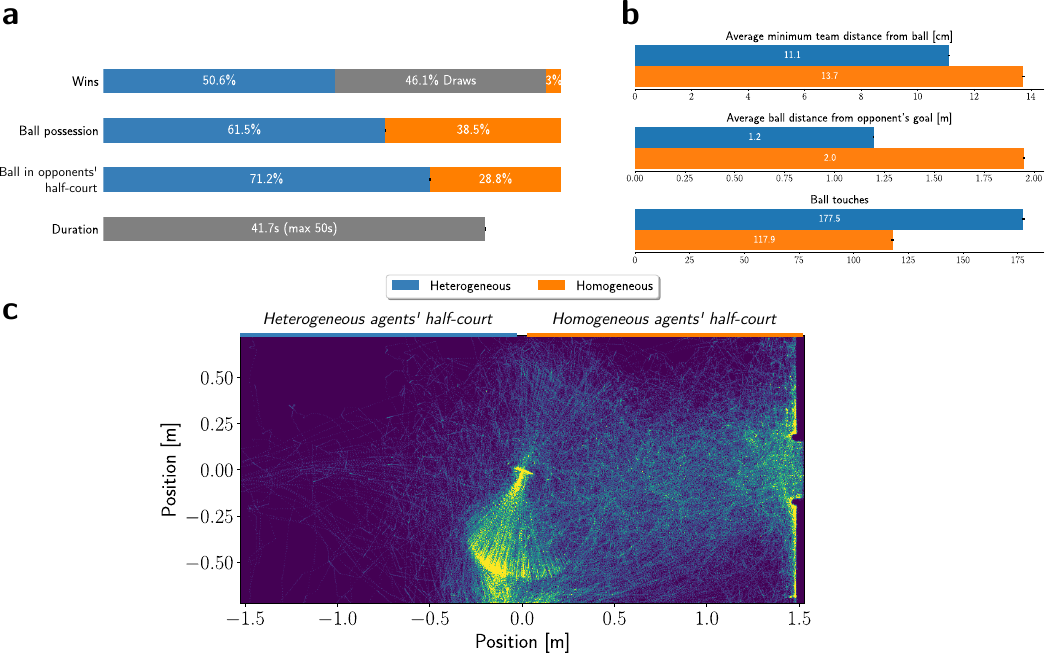}
\caption{\textbf{Results in the heterogeneous vs homogeneous \textit{Soccer} match (5v5).} The best trained models for heterogeneous and homogeneous agents in the five vs. five setting are evaluated in a competition against each other over 10,000 matches. \textbf{a,} Global statistics over the matches. \textbf{b,} Team-specific statistics over the matches. \textbf{c,} Pitch heatmap displaying the frequency of ball positions over 1,000 matches.
The results show the higher performance of heterogeneous agents, able to win over half the matches played, with higher attack and ball-handling statistics.
The heatmap corroborates these results by showing a major ball presence in the homogeneous agents' defensive half-pitch with a spatially-spread attack pattern for heterogeneous agents. On the other hand, homogeneous agents perform less frequent and more localized offensive actions.
Statistics report mean and standard error over 10,000 matches. Ball possession is computed by considering the team membership of the closest agent to the ball.
}
\label{fig:match}
\end{figure*}

\subsubsection{The final match: heterogeneous vs homogeneous teams}
\label{sec:final_match}

To further elucidate the benefits of diversity in team-play, we evaluate the best trained 5v5 models for heterogeneous and homogeneous teams in a match against each other. 
The homogeneous ($\mathrm{SND}_\mathrm{des}=0$) and heterogeneous ($\mathrm{SND}_\mathrm{des}=0.2$) models are trained in an identical setting using the same curriculum, and both obtain the maximum final reward of $\approx1$ (scoring almost all the time against the heuristic).
This evaluation quantifies their resilience to an out-of-distribution opponent and, thus, the zero-shot generalization of their strategies.
We run the evaluation over 10,000 matches, reporting various 
soccer statistics in \autoref{fig:match}ab.
The results show stronger performance for diverse teams, which win $50.6\%$ of the time, with higher ball possession, higher presence in the opponents' half-pitch, and higher interaction with the ball.
The heatmap of ball positions in \autoref{fig:match}c corroborates these results by showing a majority of ball presence in the homogeneous agents’ defensive half-pitch with a spatially spread-out attack pattern for heterogeneous agents. On the other hand, homogeneous agents perform less frequent and more localized offensive actions\footnote{Note that the high visitation sections in the center of the pitch are due to the first contact between teams and the ball, repetitively resulting in the ball bouncing in the `down' direction, and are not relevant for our insights.}.
The difference in performance during deployment against this previously unseen opponent can be explained by the fact that the homogeneous agents
only observe data collected with a homogeneous strategy, 
\textbf{while heterogeneous agents experience higher diversity in the training data, implicitly making them more resilient to unseen opponents.}

\begin{figure*}[!pt]
\centering
\includegraphics[width=\textwidth]{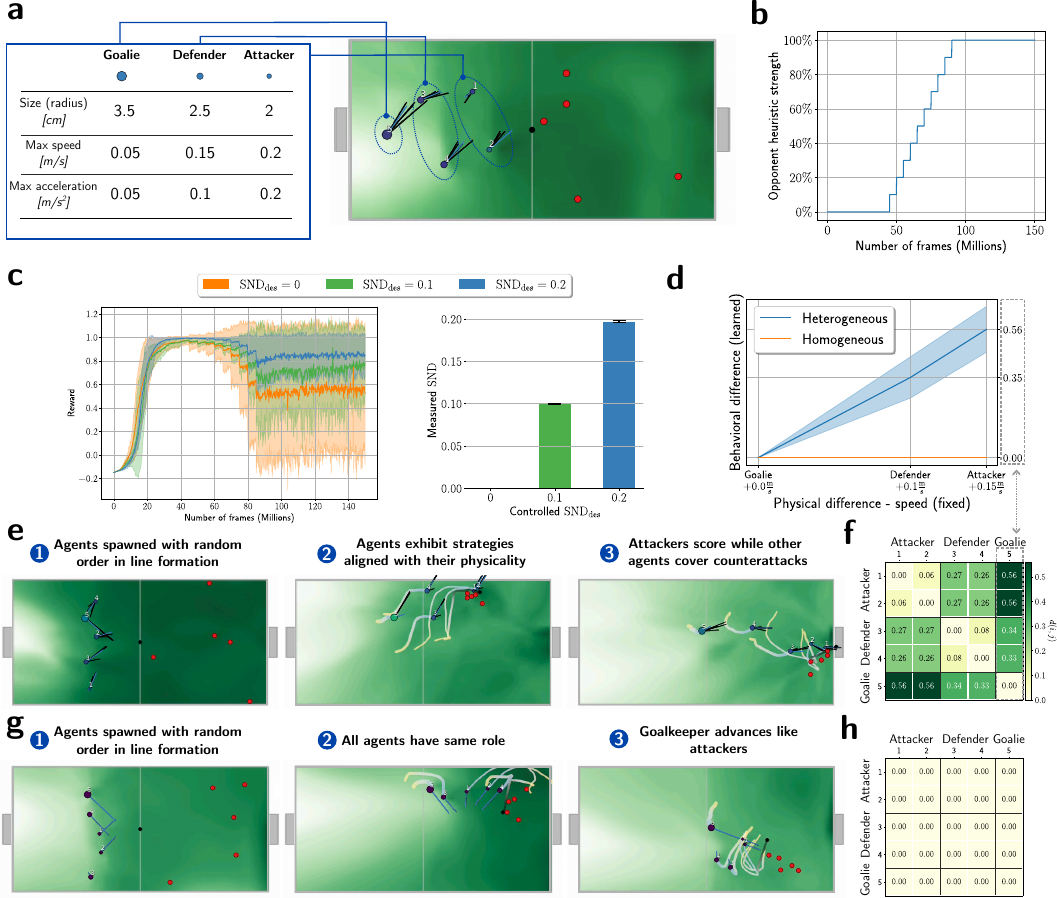}
\caption{
\textbf{Results in the physically-different 5v5 \textit{Soccer} experiment.} 
\textbf{a,} Task setup with the three types of embodiment used: \textit{Goalie} (big and slow), \textit{Defender} (average speed and size), and \textit{Attacker} (small and fast).
\textbf{b,} Curriculum used for opponents' strength annealing throughout training. 
\textbf{c,} Reward (throughout training, with $1$ obtained when agents score in $100\%$ of the matches) and diversity ($\mathrm{SND}$, after training). We report results for agents constrained at different diversity levels $\mathrm{SND}_\mathrm{des}$.
The results show that a higher diversity target leads to better performance, allowing the agents to leverage their physical differences at the behavioral level.
\textbf{d,} Learned behavioral differences (from \textit{Goalie} to other roles) as a function of fixed physical differences, showing that heterogeneous agents learn diverse behavioral roles proportional to their physical differences. Behavioral differences are evaluated over 2.5 million observations for the $\mathrm{SND}_\mathrm{des}=0.2$ model.
\textbf{e,g,} Heterogeneous (\textbf{e}) and homogeneous (\textbf{g}) agents, trained with the starting 1-2-2 formation, are evaluated starting in a random order with a line formation. Diverse agents are unaffected by this shift and are able to keep leveraging their physical differences at the behavioral level, while homogeneous agents, now not able to condition on the starting position to infer the role, exhibit the same strategy for all agents. 
The behavioral distance matrices (\textbf{f,h}) further show how heterogeneous agents are the only ones to learn behavioral differences that are proportional to the physical differences: with \textit{Goalie} and \textit{Attacker} being the furthest apart and agents with the same embodiment being behaviorally similar.
Each datapoint in the reward curves and bar charts is computed over 480 matches ($\times 3$ training seeds). Each match lasts max 500 steps or until a team scores. Reward curves report mean and standard deviation, while bar charts report mean with error bars representing the \nth{25} and \nth{75} percentiles.
}
\label{fig:phy_diff}
\end{figure*}

\subsubsection{The synergy of behavioral diversity and physical differences}
\label{sec:phy_diff}
Physical heterogeneity is commonplace in both natural and engineered collectives, yet its relation to behavioral diversity is not clear. To better understand this interplay, we design a variation of the 5v5 \textit{Soccer} task (\autoref{fig:phy_diff}a), where three types of embodiment are used: \textit{Goalie} (big and slow), \textit{Defender} (average speed and size), and \textit{Attacker} (small and fast). We train teams with different diversity targets, including homogeneous agents ($\mathrm{SND}_\mathrm{des}=0$), against the heuristic opponents with the strength curriculum shown in \autoref{fig:phy_diff}b. The results in \autoref{fig:phy_diff}c confirm that higher diversity in this task improves performance, better adapting to the increase of the opponents' strength.
To explain these results, we plot behavioral differences (\autoref{fig:phy_diff}d), learned by heterogeneous agents ($\mathrm{SND}_\mathrm{des}=0.2$), as a function of fixed physical differences (speed), from the perspective of the \textit{Goalie}. This shows that the learned behavioral distance between agents is proportional to their fixed physical difference, demonstrating that the diversity in the emergent roles aligns with the fixed physical embodiment.
To further show the benefits of this mind-body synergy, we evaluate heterogeneous ($\mathrm{SND}_\mathrm{des}=0.2$, \autoref{fig:phy_diff}e) and homogeneous ($\mathrm{SND}_\mathrm{des}=0$, \autoref{fig:phy_diff}g) agents, trained starting in a 1-2-2 formation, in a novel setting starting in a line formation with random agent order.
Diverse agents are unaffected by this change of configuration, and continue to leverage their physical differences at the behavioral level: no matter where the agent is spawned, it maintains its role.
Homogeneous agents, which were implicitly inferring their role from starting positions, are not able to adapt to this change, resulting in the same strategy being used for all agents, regardless of their embodiment. 
The behavioral matrices in \autoref{fig:phy_diff}fh confirm this, showing how heterogeneous agents are the only ones to learn behavioral differences that are proportional to physical differences. We highlight that: (1) the \textit{Goalie} and \textit{Attacker} are the furthest apart, (2) agents with the same embodiment are behaviorally similar, and (3) the \textit{Attacker}-\textit{Defender} pair is closer than \textit{Defender}-\textit{Goalie} pair, underlining the higher physical difference between the embodiments of the latter pair.

\subsection{Behavioral diversity boosts exploration}
\begin{figure*}[!pt]
\centering
\includegraphics[width=\textwidth]{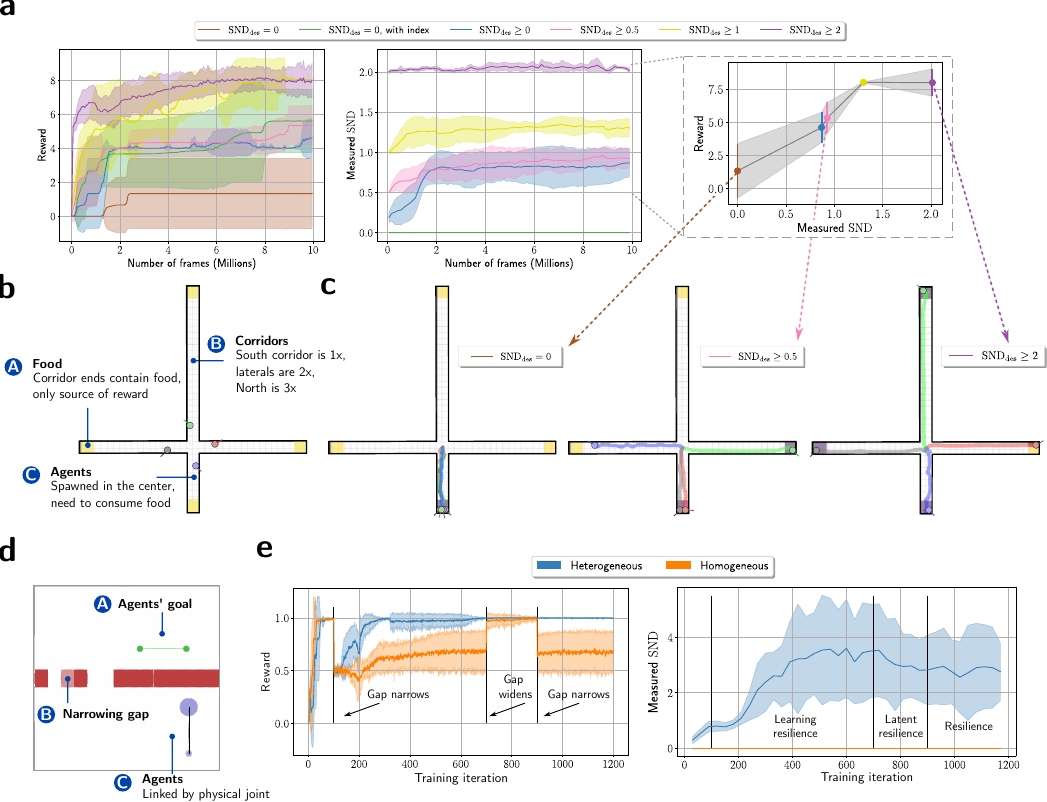}
\caption{
\textbf{Results for the exploration (a,b,c) and resilience (d,e) experiments.} 
\textbf{a,} Reward and measured diversity ($\mathrm{SND}$) in the \textit{Pac-Men} task for different diversity target ranges. We also report results for homogeneous agents ($\mathrm{SND}_\mathrm{des}=0$) with and without the addition of a one-hot index to the their observations.
The addition of this index enables homogeneous agents to condition upon it to learn multiple behavioral roles. However, such a paradigm still does not reach the performance and convergence speed of constrained heterogeneous agents. As shown in the rightmost plot, depicting reward as a function of diversity, only agents constrained to high diversity ranges are able to explore the map entirely and find all the food. The plots report mean and standard deviation over 6 training seeds.
\textbf{b,} \textit{Pac-Men} task setup.
\textbf{c,} Example renderings for different diversity levels, showing the correlation between diversity and exploration in the task. Only agents with high diversity are able to discover the food at the end of the north corridor.
\textbf{d,} \textit{Dynamic Passage} task setup.
\textbf{e,} Reward and measured diversity for homogeneous and unconstrained heterogeneous agents in the \textit{Dynamic Passage} task. These plots show the emergence of latent resilience for heterogeneous learning. Heterogeneous agents are able to acquire resilience skills when facing
a disturbance (narrowing gap) and utilize those skills when the disturbance reappears. We report mean and standard deviation over 6 random seeds for each
experiment. Each training iteration consists of 60,000 frames.
}
\label{fig:explo_res}
\end{figure*}

Exploration is key to learning from experience
, with numerous methods proposing to improve it in multi-agent learning
~\cite{mahajan2019maven,Wang*2020Influence-Based,pmlr-v139-liu21j,NEURIPS2021_1e8ca836}.
Here we show how controlling diversity intrinsically boosts exploration, thus playing an key role in the overall learning process.

To this end, we customize the \textit{Pac-Men} task~\cite{chenghao2021celebrating} (\autoref{fig:explo_res}b) so that exploration during learning becomes crucial for performance.
Four agents are spawned at the center of a four-way intersection with corridors of different lengths ($\mathrm{down}:\mathrm{left}:\mathrm{up}:\mathrm{right} = 1 : 2 : 3 : 2$). They are only able to observe a local area around them and are collectively and sparsely rewarded when consuming food at the end of the corridors. In this task, reward is proportional to exploration success as, to obtain the best performance, agents are required to venture in different directions, some taking longer paths than others.

We report results (\autoref{fig:explo_res}a) for agents controlled to reach specified diversity ranges. Unlike previous experiments, we employ an inequality constraint on the diversity target, fixing only the minimum, while leaving the maximum unbounded. We also report results for homogeneous agents ($\mathrm{SND}_\mathrm{des}=0$) with and without the addition of a one-hot index to the their observations.
The addition of this index enables homogeneous agents to condition upon it to learn multiple behaviors, avoiding the otherwise only possible policy of all going to the same food particle. However, such a paradigm still does not reach the performance and convergence speed of diversity-controlled heterogeneous agents. 
Agents controlled to reach higher diversity are able not only to achieve higher rewards and thus explore better, but also present a much faster convergence speed (as can be noticed by looking at the initial reward slope of $\mathrm{SND}_\mathrm{des}\geq2$).
The renderings in \autoref{fig:explo_res}c
further elucidate the correlation between diversity and exploration in the task, demonstrating that \textbf{a minimum amount of diversity is \textit{necessary} to achieve the full task} (i.e., \textit{all four} food sources are discovered).

\subsection{Behavioral diversity enables resilience}

Diversity in natural systems has been shown to 
enable resilience to environmental disruptions~\cite{https://doi.org/10.1111/j.1365-2664.2011.02048.x}. In this section, we demonstrate that a similar paradigm emerges in artificial collectives that exhibit behavioral diversity.

We create the \textit{Dynamic Passage} task (\autoref{fig:explo_res}d).
Two different-sized agents, physically linked by a joint, need to traverse horizontally two randomly-spawned gaps to reach their goal. While both gaps are initially big enough to fit either agent, the task undergoes a disruption, consisting in one of the gaps narrowing, such that it impedes passage to the lager agent.
To overcome this disturbance, agents need to cooperate to find their fitting passageway.

We train unconstrained heterogeneous and homogeneous agents, reporting reward and diversity in \autoref{fig:explo_res}e.
There is no disturbance through iterations 0 to 100, and thus both paradigms learn to solve the task with resulting behavioral homogeneity (heterogeneous agents converge to low diversity values).
At iteration 100, the gap narrows, causing both paradigms drop to a reward of $0.5$, succeeding the task approximately $50\%$ of the time.
During iterations 100-700, heterogeneous agents increase their diversity and learn how to overcome the disturbance, effectively assigning the bigger agents to the bigger gap---a strategy that homogeneous agents are incapable of. 
At iteration 700, the gap is widened again, enabling both paradigms to regain the original performance.
However, the high $\mathrm{SND}$ metric suggests that \textbf{the heterogeneous model has learned a \textit{latent resilient} skill}.
This is confirmed when the same disturbance is reintroduced (iteration 900), having no effect on the performance of heterogeneous agents, while the homogeneous counterpart suffers the same impairment as before.

\section{Discussion}\label{sec:dicsussion}

In this work, we studied the impact and benefits of behavioral diversity in multi-agent reinforcement learning.
To do this, we decomposed collective learning into three core challenges: cooperation (team play), exploration, and resilience.
For each, we showed how heterogeneity provides non-trivial benefits over homogeneous training, demonstrating that diversity is core in all three aspects and, thus, in collective learning as a whole.

The insights gathered in this study constitute important progress in the overarching goal of understanding and developing collective machine intelligence.
In particular, in \textit{Soccer}, we have shown the emergence of unbiased behavioral roles in the form of \textit{passing} and \textit{goalkeeping}: known strategies already present in human teams that are now confirmed to naturally emerge and outperform spatially-concentrated homogeneous strategies.
Furthermore, this emergence occurs and is beneficial even in the case of physically-identical agents, an insight that is hard to gather from human teams.
In the case of physically-different players, we showed how they are able to learn diverse behavioral roles corresponding to their embodiment, and leverage this mind-body alignment for increased performance and resilience.
We also demonstrated how constraining behavioral diversity acts as a natural exploration enabler, even in the absence of any explicit exploration objective or reward. Lastly, we saw how diverse agents collectively acquire and maintain latent skills that enable resilience, and leverage them to overcome repeated disruptions throughout learning.

In future work, we are interested in extending our study to a continual learning setting on real-world robot platforms, investigating the role of diversity in lifelong learning for physical agents.

\section{Methods}\label{sec:methods}

\subsection{Multi-agent reinforcement learning (MARL) preliminaries}

Reinforcement Learning (RL) is a paradigm in which agents learn to interact with an environment to maximize a given reward signal.
MARL is an extension of RL that involves multiple agents. Our MARL setup can be formalized as a partially observable Markov game.

\textbf{Partially Observable Markov Games.} A Partially Observable Markov Game (POMG)~\cite{shapley1953stochastic} is defined as a tuple
$$\left \langle \mathcal{N}, \mathcal{S}, \left \{ \mathcal{O}_i \right \}_{i \in \mathcal{N}}, \left \{ \sigma_i \right \}_{i \in \mathcal{N}},  \left \{ \mathcal{A}_i \right \}_{i \in \mathcal{N}}, \left \{ \mathcal{R}_i \right \}_{i \in \mathcal{N}}, \mathcal{T}, \gamma \right \rangle,$$
where $\mathcal{N} = \{1,\ldots, n\}$ denotes the set of agents,
$\mathcal{S}$ is the state space, and,
$\left \{ \mathcal{O}_i \right \}_{i \in \mathcal{N}}$ and
$\left \{ \mathcal{A}_i \right \}_{i \in \mathcal{N}}$
are the observation and action spaces, with $\mathcal{O}_i \subseteq \mathcal{S}, \; \forall i \in \mathcal{N}$. 
Further, $\left \{ \sigma_i \right \}_{i \in \mathcal{N}}$ 
and
$\left \{ \mathcal{R}_i \right \}_{i \in \mathcal{N}}$
are the agent observation and reward functions (potentially identical for all agents\footnote{In this work we consider problems represented as Dec-POMDPs~\cite{bernstein2002complexity}, which are a particular subclass of POMGs with one shared reward function.}), such that
$\sigma_i : \mathcal{S} \mapsto \mathcal{O}_i$, and,
$\mathcal{R}_i: \mathcal{S} \times \left \{ \mathcal{A}_i \right \}_{i \in \mathcal{N}} \times \mathcal{S} \mapsto \R$.
$\mathcal{T}$ is the stochastic state transition model, defined as $\mathcal{T} : \mathcal{S} \times \left \{ \mathcal{A}_i \right \}_{i \in \mathcal{N}}   \mapsto  \Delta\mathcal{S}$, which outputs the probability $\mathcal{T}(s^t, \left \{ a^t_i \right \}_{i \in \mathcal{N}},s^{t+1})$ of transitioning to state $s^{t+1} \in \mathcal{S}$ given the current state $s^t \in \mathcal{S}$ and actions $\left \{ a^t_i \right \}_{i \in \mathcal{N}}$, with $a^t_i \in \mathcal{A}_i$. $\gamma \in (0,1]$ is the discount factor.

We structure the agents in a communication graph $\mathcal{G} = \left (\mathcal{N},\mathcal{E} \right )$. Nodes $i \in \mathcal{N}$ represent agents and edges $e_{ij} \in \mathcal{E}$ represent communication links.  The communication neighborhood of each agent is defined as $\mathcal{N}_i \equiv \{v_j \, | \, e_{ij} \in \mathcal{E}\}$.

At each timestep $t$, each agent $i$ gets an observation $o_i^t = \sigma_i  (s^t  ) \in \mathcal{O}_i$ that is a portion of the global state $s^t \in \mathcal{S}$.
A stochastic policy $\pi_i$, which can optionally involve communication with neighbors, uses this information to compute an action $a_i^t \sim \pi_i  (\cdot \rvert o_i^t  )$. The agents' actions $\mathbf{a}^t =  (a_1^t,\ldots,a_n^t  ) \in \mathcal{A}$, along with the current state $s^t$, are then used in the transition model to obtain the next state $s^{t+1} \sim \mathcal{T} \left (\cdot \rvert s^t,\mathbf{a}^t \right )$. A reward $r_i^t = \mathcal{R}_i  \left (s^t,\mathbf{a}^t,s^{t+1} \right )$ is then fed to agent $i$. 

The goal of each agent is to learn the policy $\pi_i$ to maximize the sum of discounted rewards $ v_i^t = \sum_{k=0}^{T} \gamma^k r_{i}^{t+k}  $ over an episode with horizon $T$, potentially infinite\footnote{$\gamma^k$ indicates $\gamma$ to the power of $k$, and not the timestep superscript.}. $v_i^t$ is called the return. 
We employ learned value functions to estimate the expected return starting from a given observation and following the current policy.

\subsection{Measuring diversity}
We employ System Neural Diversity ($\mathrm{SND}$)~\cite{bettini2023snd} to measure the diversity of policies $\pi_i$. This metric is briefly presented in the introduction and explained in depth in the respective paper.

In relation to this work, we highlight a few properties of this metric that are leveraged. In particular, we parameterize actions as multivariate normal distributions, independent in each action dimension. This allows to compute the Wasserstein distance $W_2$ between agent pairs in a closed-form.
\begin{definition}[Wasserstein metric for multivariate normal distributions]
Let $\pi_1 = \mathcal{N}(\mu_1,\Sigma_1)$ and $\pi_2 = \mathcal{N}(\mu_2,\Sigma_2)$ be two multivariate normal distributions on $\R^m$. Then, the 2-Wasserstein distance between $\pi_1$ and $\pi_2$ is computed as:
\begin{equation*}
W_2(\pi_1,\pi_2) = \sqrt{||\mu_1-\mu_2||^2_2  + \mathrm{trace}(\Sigma_1+\Sigma_2- 2(\Sigma_2^{\frac{1}{2}}\Sigma_1\Sigma_2^{\frac{1}{2}})^{\frac{1}{2}})} .
\end{equation*}
\end{definition}
We consider policies with the form $\pi_i(o) = \mathcal{N}(\mu_i(o),\sigma_i(o))$, with $\mu_i(o),\sigma_i(o)\in\R^m$, where $\sigma_i$ is a standard deviation vector which uniquely defines a diagonal covariance matrix $\Sigma_i(o)\in\R^{m\times m}$, $\Sigma_i(o) = \mathrm{diag}(\sigma_i(o)^2)$.

We further leverage the ``Invariance in the number of behaviorally equidistant agents'' property of $\mathrm{SND}$, which implies that the value of the metric remains constant when varying the number of agents in the system at the same behavioral distance. This property allows comparing diversity across systems of different size.

\subsection{Controlling diversity}

We employ Diversity Control (DiCo)~\cite{bettini2024controlling} to control the diversity of policies $\pi_i$. This paradigm is briefly presented in the introduction and explained in depth in the respective paper.

We update the original implementation with additional features. We enable gradient propagation through the computation of the scaling factor to improve training stability. We extend the method to allow inequality diversity constraints (e.g., $\mathrm{SND}_\mathrm{des}\geq x$). These are used for the exploration experiments in \autoref{fig:explo_res}a-c. Inequality constraints are achieved by rescaling the policies only in case the current diversity falls outside of the desired range.
For the \textit{Soccer} 5vs5 tasks, we add shared encoder networks (i.e., Deep Sets, GNN) before the DiCo models to compute a context encoding, as shown in \autoref{fig:5v5}b. This improves training efficiency by allowing the diverse DiCo networks to condition of this context instead of raw observations and thus avoid extra computation that is better suited to homogeneous models.

\subsection{Training details}

Training is performed in the BenchMARL library~\cite{bettini2024benchmarl} using the TorchRL framework~\cite{bou2023torchrl}. Tasks are implemented and open-sourced in the Vectorized Multi-Agent Simulator (VMAS)~\cite{bettini2022vmas}. 
All experiments are performed using the Proximal policy Optimization (PPO) training algorithm~\cite{schulman2017proximal}, either applied independently to all agents (IPPO~\cite{de2020independent}, used in \textit{Pac-Men} and \textit{Tag}), or with a critic that takes all agents' data as input (MAPPO~\cite{yu2022surprising}, used in \textit{Soccer} and \textit{Dynamic passage}).
Experiment configurations leverage Hydra~\cite{Yadan2019Hydra} to decouple YAML configuration files from the Python codebase.
The attached code contains thorough instructions on the meaning of configuration parameters and reproducing instructions.
Neural network models used for policies (actors) and  value functions (critics) are Multi Layer Perceptrons (MLPs). In the case of 2vs2 \textit{Soccer} and \textit{Dynamic passage}, policies are able to access data from the other agent (e.g., via communication). In the case of \textit{Pac-Men} and \textit{Tag}, policies only have access to data from the ego agent.

\subsubsection{5vs5 policy model}
The policy model used in the \textit{Soccer} 5vs5 tasks has a more complex architecture than simple MLPs to deal with the high (and possibly variable) number of agents and adversaries.
The architecture, shown in \autoref{fig:5v5}b, has a first homogeneous encoding layer to process adversary data using a Deep Sets~\cite{deepsets} model:
\begin{equation*}
    h_i = \rho_\theta\left (o_i \| \sum_{j\in \mathrm{Opp}} \phi_\theta(o_j)\right ),
\end{equation*}
meaning that the hidden state $h_i\in\R^d$ of agent $i$ is computed by applying a permutation-invariant operation on the opponents' observations $\{o_j\}_{j\in\mathrm{Opp}}$ and processing the result, concatenated with the agent's observations, using an MLP.
These hidden states, alongside ball and agent information, are used as node features in a Graph Neural Network (GNN), enabling agents to communicate.
The absolute position ${p}_i \in \R^2$ and the agent velocity ${v}_i \in \R^2$ are used to compute edge features $e_{i,j}$, which are relative features of agents $i$ and $j$. In this work, we use the relative position ${p}_{i,j} = {p}_i - {p}_j$, distance $\|p_{i,j}\|$ and relative velocity ${v}_{ij} = {v}_i - {v}_j$ as edge features. The edge features $e_{ij}$ and the agent hidden state $h_i$ are then processed using a Graph Attention Network (GAT)~\cite{brody2022how}:
\begin{equation*}
        z_i = \sum_{j \in \mathcal{N}_i \cup \{ i \}}
        \alpha_{i,j,\theta}(h_i,h_j,e_{i,j}) \psi_\theta (h_{j}),
\end{equation*}
where $\alpha_{i,j,\theta}(h_i,h_j,e_{i,j})$ is the attention coefficient that agent $i$ assigns to the hidden state of agent $j$, processed by an MLP $\psi$.
All the networks presented so fare are homogeneous and parameterized by $\theta$, shared for all agents. The encoding $z_i$ is used as the context and is fed to the DiCo models to compute the homogeneous policy component $\pi_h(z_i) = \zeta_\theta(z_i)$ and the heterogeneous deviations $\pi_{h,i}(z_i) = \chi_{\eta_i}(z_i)$. $\zeta$ and $\chi$ are MLPs with shared parameters $\theta$ and per-agent heterogeneous parameters $\eta_i$ respectively.

\backmatter

\section*{Declarations}
\subsection*{Supplementary information}

Supplementary information, tables, and figures are reported in the Appendix of this file.

\subsection*{Data availability}
All raw experiment data used in our results is available upon request and will be uploaded to Zenodo upon publication.

\subsection*{Materials and code availability}
The documented code to reproduce the results reported in this manuscript 
is available upon request and will be made be publicly available on GitHub upon publication.


\subsection*{Acknowledgements}
This work was supported by Army Research Laboratory (ARL) Distributed and Collaborative Intelligent Systems and Technology (DCIST) Collaborative Research Alliance (CRA) W911NF-17-2-0181 and the European Research Council (ERC) Project 949940 (gAIa).


\subsection*{Author contributions}
Matteo Bettini performed the research, wrote the code, and analyzed the results. Matteo Bettini and Amanda Prorok designed the research. Ryan Kortvelesy and Matteo Bettini implemented the heuristic algorithm for the \textit{Soccer} opponents. Amanda Prorok initiated and supervised the project, and acquired the funding. Matteo Bettini, Ryan Kortvelesy, and Amanda Prorok wrote the paper.

\subsection*{Competing interests}
The authors declare no competing interests.

\newpage
\begin{appendices}


\section{Computational resources used}

For the realization of this work, several hours of compute resources have been used.
In particular, they have gone towards: experiment design, prototyping, and running final experiments results.
Simulation and training are both run on GPUs, so no CPU compute has been used.
Among the tasks, \textit{Soccer} has required the majority of  compute due to the long training times required deriving from its complexity.
We estimate:
\begin{itemize}
    \item  500 compute hours on an NVIDIA GeForce RTX 2080 Ti GPU 
    \item  1500 compute hours on am NVIDIA L40S GPU
    \item  5000 compute hours on an NVIDIA A100-SXM-80GB GPU
\end{itemize}

\section{Additional \textit{Soccer} results}

\subsection{Further heterogeneous emergent strategies}

In this section, we analyze further strategies that emerge when constraining agents to be heterogeneous. The heterogeneous constraints indirectly improve the team's spatial coverage of the pitch, playing a similar role to that of a soccer coach. However, they do not prescribe any specific behavior and do not bias the strategies with human knowledge, making all the results shown purely emergent.

In the following, we analyze some of the emergent heterogeneous strategies in the two vs. two (\autoref{sec:het_2v2}) and five vs. five (\autoref{sec:het_5v5}) settings.

\subsubsection{Two vs. two}
\label{sec:het_2v2}
\begin{figure*}[!tp]
\centering
\includegraphics[width=\textwidth]{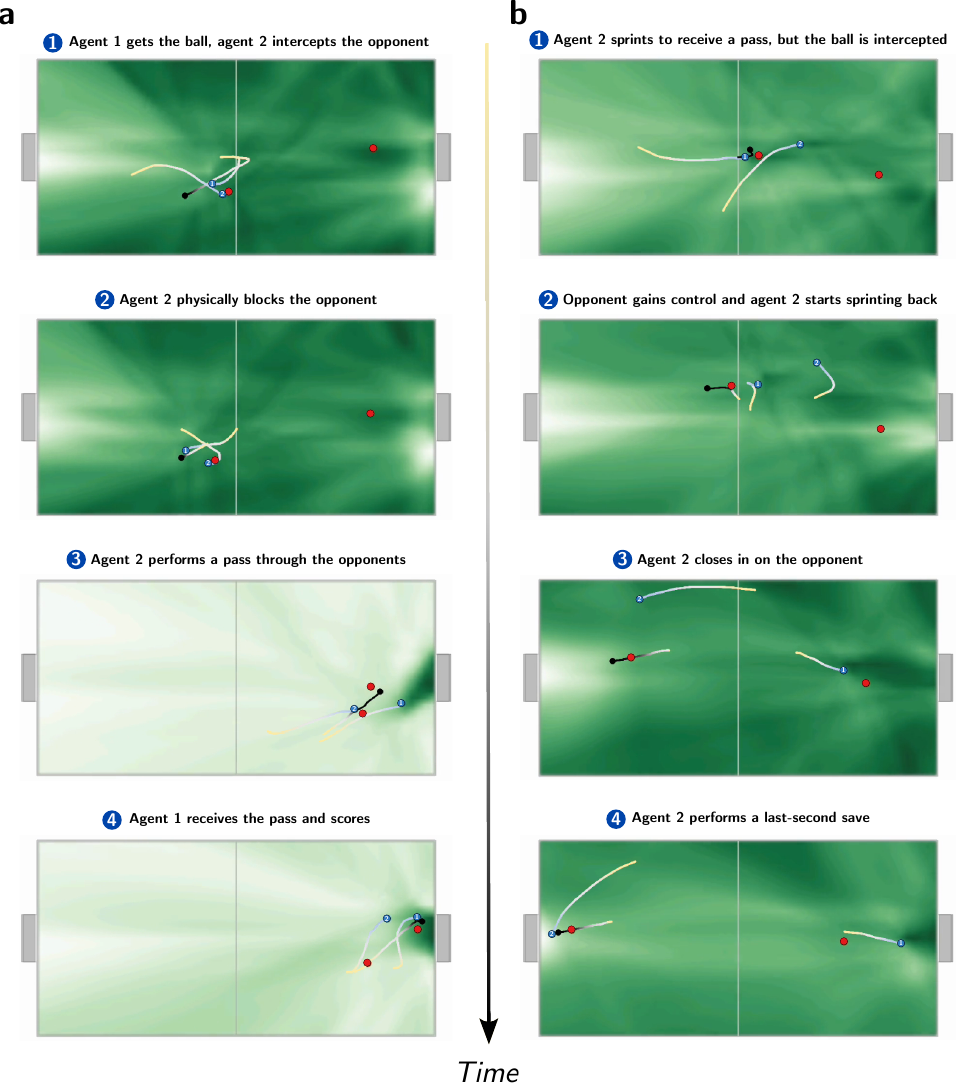}
\caption{\textbf{Further emergent heterogeneous strategies in the two vs. two setting}. \textbf{a,} ``opponent socking'' strategy where one agent focuses on physically blocking the opponent, while the other agent goes for the ball undisturbed. 
\textbf{b,} ``last-second save'' strategy where one agent is able to intercept an opponent about to score from the other side of the pitch.
}
\label{fig:het_strats_2v2}
\end{figure*}

 In \autoref{fig:het_strats_2v2} we show two further strategies that emerge when training diverse agents in the 2v2 setting.

 \autoref{fig:het_strats_2v2}a shows an ``opponent blocking'' strategy. Here, in the initial moments of the match, agent 2 focuses on physically blocking the opponent, while agent 1 goes for the ball undisturbed. This proves to be a successful strategy as it is able to buy precious time for agent 1\footnote{Note that physical collisions are not penalized}. The agents then proceed to score thanks to a pass from agent 2 through the opponents.

 \autoref{fig:het_strats_2v2}b shows a ``last-second save'' strategy. Here, as soon as the agents loose control of the ball, agent 2 notices it and starts rushing back. By rushing on the side of the pitch, it is able to intercepts the opponent, which was otherwise proceeding undisturbed, and avoid a sure goal at the last second.

It is important to note that in both these scenarios only one agent chases the ball at a time, a positive characteristic that we hardly see emerging from homogeneous agents.

\subsubsection{Five vs. five}
\label{sec:het_5v5}
\begin{figure*}[!tp]
\centering
\includegraphics[width=\textwidth]{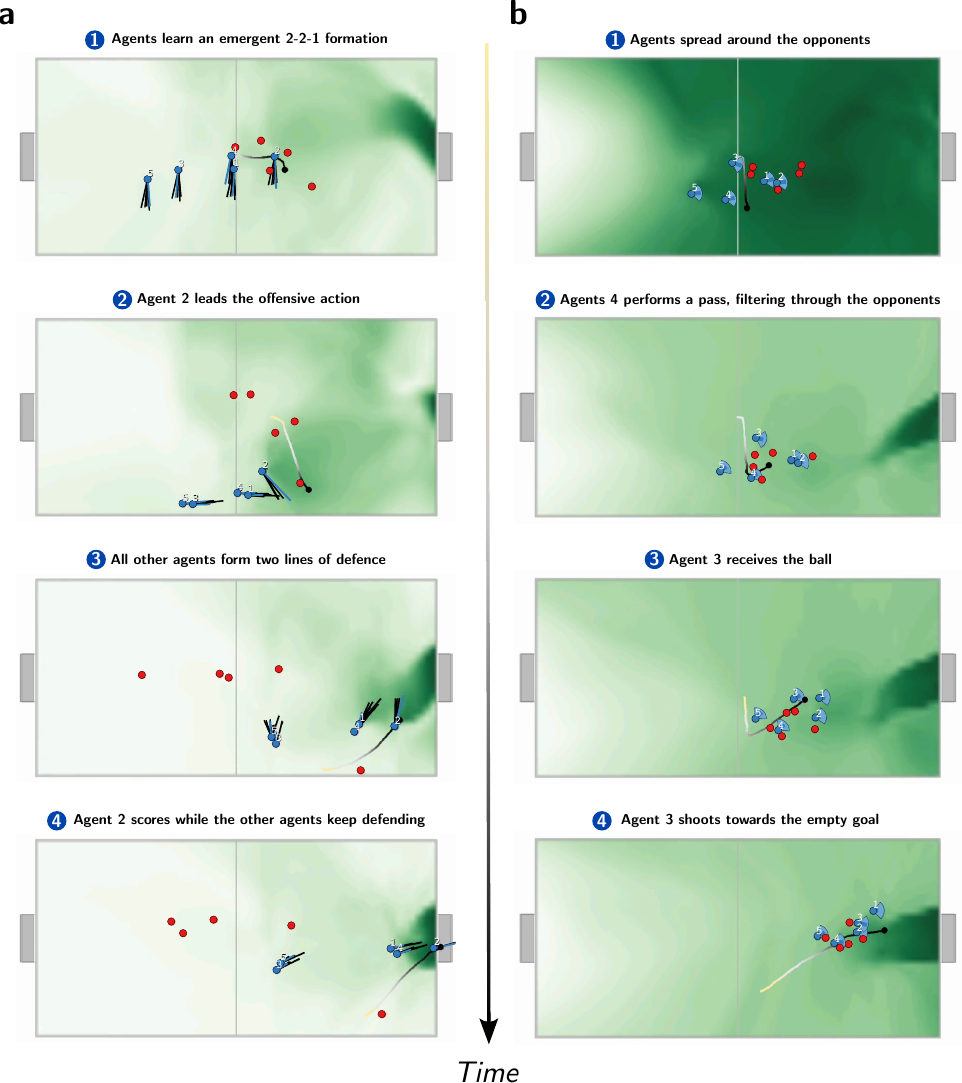}
\caption{\textbf{Further emergent heterogeneous strategies in the five vs. five setting}.
}
\label{fig:het_strats_5vs5}
\end{figure*}

In \autoref{fig:het_strats_5vs5} we show two further strategies that emerge when training diverse agents in the 5vs5 setting.

\autoref{fig:het_strats_5vs5}a shows the emergence of a 2-2-1 formation. Here, agent 2 leads the attack, while all other agents stay further back, forming two lines of defense. This strategy proves resilient in case agent 2 loses control of the ball, as the other agents can still intercept the opponents' counterattack.

\autoref{fig:het_strats_5vs5}b shows the emergence of a through pass for \textit{kicking} agents. Here, the agents position themselves around the opponents' team. Agent 4 is then able to perform a pass, which goes through the opponents and reaches agent 3, that proceeds to shoot in the empty goalpost.

Again, both these strategies show good spatial coverage and little cluttering around the ball, features that are rarely obtained from homogeneous agents.

\subsection{Homogeneous strategies}

Having analyzed the emergent strategies of heterogeneous agents, we now turn our attention to the strategies learned by homogeneous ones.
In particular, we highlight how these strategies prove myopic and resemble the ones emerging in human teams that approach the sport without coaching.

In all task setups, homogeneous agents learn to converge and group around the ball. The is due to the fact that ball movement is the main source of reward. Collectively converging towards the reward source can appear as the best way to obtain reward, but shows its limitations in the long term.
In fact, despite this strategy leading to success in certain cases (due to the overpowering physical superiority of a large group) it shows significant long-term limitations. Most importantly, spatial grouping prevents passing and proves brittle when opponents gain control of the ball and dribble past the group, not encountering any further line of defense.

This type of behavior also emerges in human teams, where players tend to collectively chase the ball in the first phases of learning, effectively not showing good spatial coverage of the pitch. In this analogy, our diversity constraint acts in a similar way to a football coach, incentivizing players to learn different strategies and, as a consequence, improve spatial coverage.

In the following, we analyze some of the emergent homogeneous strategies in the two vs. two (\autoref{sec:hom_2v2}) and five vs. five (\autoref{sec:hom_5v5}) settings and highlight their long-term inefficiencies.

\subsubsection{Two vs. two}
\label{sec:hom_2v2}
\begin{figure*}[!tp]
\centering
\includegraphics[width=\textwidth]{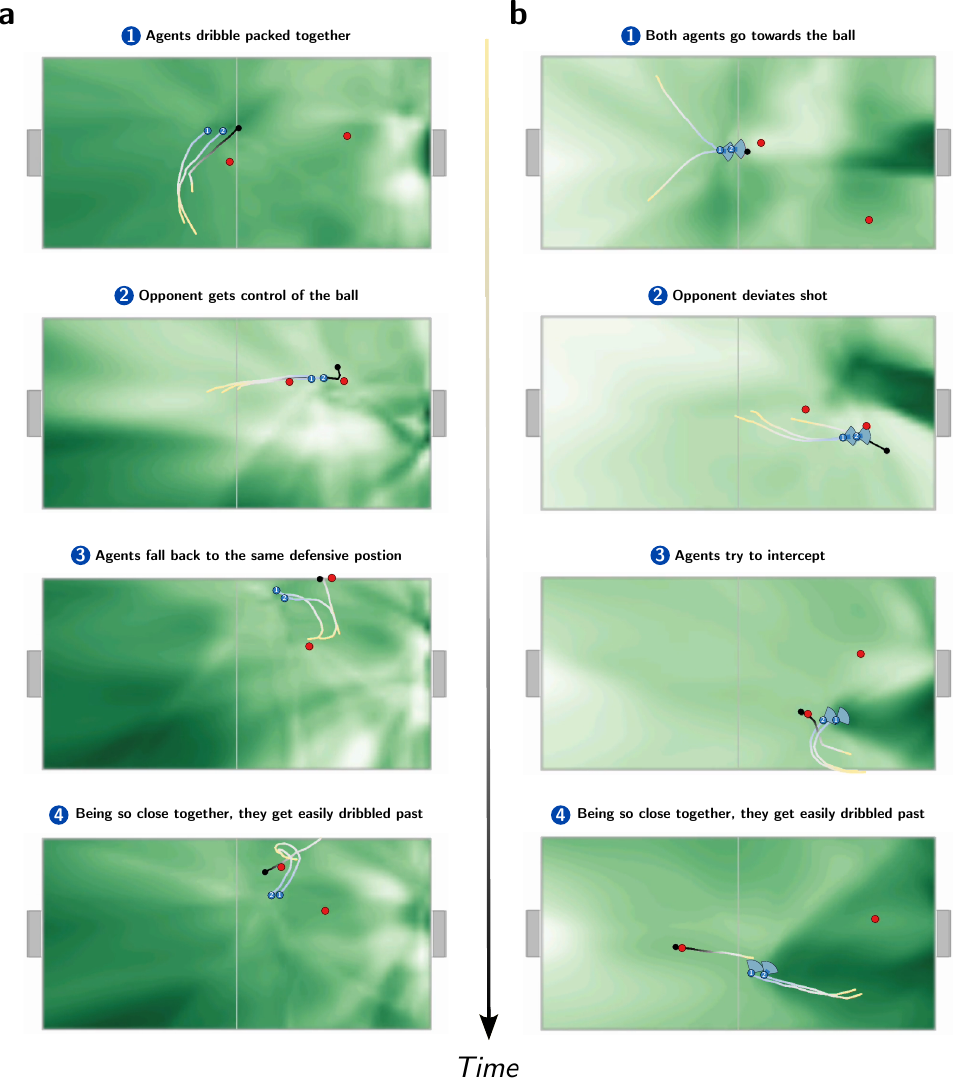}
\caption{\textbf{Homogeneous strategies in the two vs. two setting}.
\textbf{a,} \textit{Non-kicking} agents. \textbf{b,} \textit{Kicking} agents.
These renderings show that homogeneous agents tend to stay packed together, making them unable to recover once the opponents gain control of the ball.
}
\label{fig:homog_2v2}
\end{figure*}

\autoref{fig:homog_2v2} reports examples of learned homogeneous strategies for \textit{non-kicking} (\autoref{fig:homog_2v2}a) and \textit{kicking} (\autoref{fig:homog_2v2}b) agents in the 2v2 soccer task.
In both cases, the agents stay packed together to dribble the ball towards the opponents' goalpost. As soon as one of the opponents intercepts them and gains control of the ball, they try to fall back and start defending. However, being so close together, they are both too advanced, with no agent in a defensive position. Thus, the opponent is able to score undisturbed.

\subsubsection{Five vs. five}
\label{sec:hom_5v5}
\begin{figure*}[!tp]
\centering
\includegraphics[width=\textwidth]{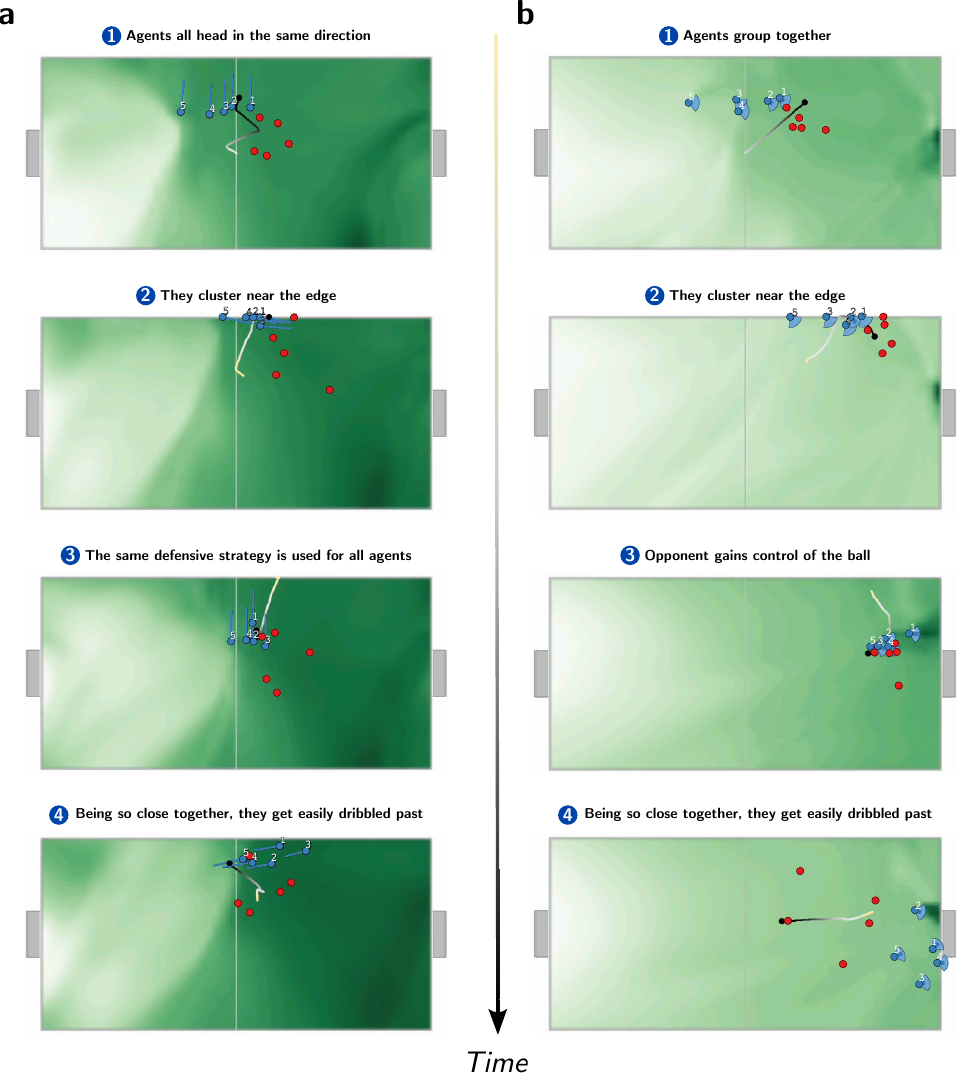}
\caption{\textbf{Homogeneous strategies in the five vs. five setting}.
\textbf{a,} \textit{Non-kicking} agents. \textbf{b,} \textit{Kicking} agents.
These renderings show that homogeneous agents tend to stay packed together, making them unable to recover once the opponents gain control of the ball.
}
\label{fig:homog_5v5}
\end{figure*}

\autoref{fig:homog_5v5} reports examples of learned homogeneous strategies for \textit{non-kicking} (\autoref{fig:homog_5v5}a) and \textit{kicking} (\autoref{fig:homog_5v5}b) agents in the 5v5 soccer task.
Also in these cases, the agents learn to act similarly, despite being able to learn different behaviors based on the context.
Their dense attack pattern can prove quite strong in advancing towards the goal, but as soon they loose control of the ball, they are not able to recover due to the absence of goalkeepers or agents playing defense.

\section{Task details}
\begin{table}[ht]
\caption{Categorization of tasks analyzed in this work.}\label{tab:tasks}
\begin{tabular*}{\textwidth}{@{\extracolsep\fill}lcccccc}
\toprule
Task & Reward\footnotemark[1] & Communication\footnotemark[2] & Physical differences\footnotemark[3] & Objective\footnotemark[4] \\
\midrule
\textit{Soccer}  & Sparse+Dense & Yes &Yes+No & Shared\\
\textit{Pac-Men}  & Sparse & No & No & Shared  \\
\textit{Dynamic Passage}  & Sparse+Dense & Yes & Yes & Shared \\
\botrule
\end{tabular*}
\footnotetext[1]{Reward structure can be sparse (only given at certain timesteps) or dense (given at all timesteps).}
\footnotetext[2]{Whether the agents use communication in the policies to access each other observations.}
\footnotetext[3]{If the agents have physical differences (e.g., maximum speed, size).}
\footnotetext[4]{If the agents have a global (shared) reward function or local independent ones.}
\end{table}

In this section, we provide a detailed description of he tasks analyzed in this work. Explaining in detail their actions, observations, rewards, and termination structure. \autoref{tab:tasks} categorizes these tasks along the following classification axis: reward structure, use of communication in the policies, physical differences between the agents, and objective alignment.

\subsection{\textit{Soccer}}

The different setups of the soccer task are shown in \autoref{fig:2v2}a, \autoref{fig:5v5}a, and \autoref{fig:phy_diff}a.
They vary in number of agents, starting locations, and embodiment type.
Learning agents (blue) are trained against opponents (red) executing a hand-designed heuristic. 
Learning agents can be spawned either (1) at uniformly random positions in their half-pitch (\autoref{fig:2v2}), (2) in a 1-2-2 formation with added uniform noise (\autoref{fig:5v5}, \autoref{fig:match}, \autoref{fig:phy_diff}abcd), and (3) in a line formation with added uniform noise with random order permutations (evaluations in \autoref{fig:phy_diff}efgh). The ball is always spawned in the center of the pitch.

\subsubsection{Observation}
Agents have full observability over teammates, opponents, and the ball.
In particular, they observe: their own action force, position, velocity, relative position to the ball, relative velocity to the ball, relative position of the ball to the opponents' goalpost, ball velocity, ball acceleration, relative position and velocity to all opponents, and the opponents' absolute velocities and accelerations. Through communication they are able to access the observations of other teammates as well as compute relative position and velocity to all teammates. Shooting agents additionally observe their rotational heading.
 
\subsubsection{Action}
We consider two setups: (1) where agents can only move holonomically with 2D continuous action forces and have to physically touch the ball to dribble, and (2) where agents have two additional actions for rotating and kicking the ball with a continuous force (given that the ball is within feasible range and distance). Only the kicking action of the agent (including opponents) closest to the ball is allowed to take effect.
Opponents are always using actions of type (1).

\subsubsection{Reward}

Agents receive sparse reward of 100 for scoring a goal (positive) and for conceding a goal (negative). 

Agents receive a dense reward for moving the ball closer to the opponents' goalpost. Calling $d_{\mathrm{ball},\mathrm{goal}}^t \in \R$ the distance between the goalpost and the ball at time $t$, this reward is computed as:
$$
10 (d_{\mathrm{ball},\mathrm{goal}}^{t-1} - d_{\mathrm{ball},\mathrm{goal}}^t), 
$$
incentivizing ball movement towards the goalpost independently of absolute positions.

Lastly, in the initial phase of learning, before the addition of the opponents, we add a small dense reward to incentivize exploration and interaction with the ball. This reward is later removed as the opponent strength gets greater than 0. Calling $d_{i_{\mathrm{close}},\mathrm{ball}}^t \in \R$ the distance between the agent closest to the ball $i_\mathrm{close}$ and the ball at time $t$, this reward is computed as:
$$
0.1 (d_{i_{\mathrm{close}},\mathrm{ball}}^{t-1} - d_{i_{\mathrm{close}},\mathrm{ball}}^t), 
$$
incentivizing at least one agent to interact and stay close to the ball. To avoid penalizing ball movements, this reward is not given when the closest agent is closer than a minimum radius around the ball or when the ball is in motion.

\subsubsection{Termination}
The match terminates when a goal is scored or after 500 steps. Each simulation step corresponds to 0.1 seconds.

\subsubsection{Opponents}

Opponents are always spawned at uniformly random positions in their half-pitch.

The soccer heuristic is a hand-crafted policy that defines the policy for a variable-sized team with tunable difficulty. It is constructed with a high-level planner and a low-level controller. 
The high-level outputs trajectories specified by an initial and terminal position and velocity. Then, the low-level controller generates a trajectory with a Hermite spline, and uses a closed-loop controller to track that trajectory. The high-level policy is split into a dribbling policy, when an agent has possession, and off-the-ball movement, when it does not. The in-possession policy generates a dribbling effect with a trajectory that ends in the same position as the ball with a terminal velocity pointing towards the goal. On the other hand, the out-of-possession policy samples random positions on the pitch and evaluates their utility with a hand-crafted value function, changing current agent destination if a better alternative is found. As input, the heuristic requires the position and velocity of the ego agent, all teammates, and the ball, as well as the positions of the goalposts. The handcrafted value function for off-the-ball movement prioritizes positions that 1) are close to the ball, 2) avoid being against a wall, 3) avoid blocking the path between the ball and the target goal, 4) block the agent’s own goal on defense, and 5) avoid being bunched up next to teammates. 

There are three difficulty parameters built in to the AI, each ranging from 0 to 1: speed strength, decision strength, and precision strength. Speed strength simply dictates the speed of all movement, from dribbling to off-the-ball movement. A value of 0 yields completely stationary agents, and a value of 1 yields a behavior that constantly operates at the maximum allowable velocity. Decision strength affects the agents’ decision-making by adding noise to the value functions that dictate the agent’s decisions. This affects both movement decisions (i.e. the agent might choose to navigate to a position with a lower value, especially if the values have similar magnitude), as well as possession decisions (i.e. the agents will not be able to accurately evaluate whose possession it is, and might consequently assign the ball to the wrong teammate). Lastly, precision strength controls the agent’s ability to execute planned maneuvers. It adds noise to the target position and velocity, resulting increasingly uncoordinated behavior.

\subsection{\textit{Pac-Men}}

The \textit{Pac-Men} task was introduced in prior work~\cite{chenghao2021celebrating}. We slightly modify the  task, keeping the same objective and concept. In this task, shown in \autoref{fig:explo_res}b, four agents are spawned at the center of a four-way intersection with corridors of different lengths ($\mathrm{down}:\mathrm{left}:\mathrm{up}:\mathrm{right} = 1 : 2 : 3 : 2$). The background is divided in cells of agent's size. The cells at the end of the corridors contain food, correspoding to reward for the team. There are no inter-agent collisions in this task.

\subsubsection{Observation}

Agents observe a local 3x3 grid around their position. Each cell in this grid has a binary value representing if it contains food. Therefore, agents will all have the same (empty) observation when they are in search of food or when they are in an area where food has already been consumed. Agents do not observe their absolute position.

\subsubsection{Action}
Agents have a 2D continuous action force that determines their movement.

\subsubsection{Reward}
All agents are collectively rewarded with a sparse reward of 1 when a food particle is consumed. Each food particle can only be consumed once, thus multiple agents are not incentivised to sample the same area.

\subsubsection{Termination}
The task is done after 300 steps.  Each simulation step corresponds to 0.1 seconds.

\subsection{\textit{Dynamic Passage}}

This task, shown in \autoref{fig:explo_res}d, involves two agents of different sizes (blue circles), connected by a rigid linkage through two revolute joints.
The team needs to cross a passage while keeping the linkage parallel to it and then match the desired goal position (green circles) on the other side.
The passage is comprised of two gaps in the wall that have the same size and can fit either of the agents. The gaps are spawned in a random position and order on the wall, but always at the same distance between each other.
The team is spawned in a random order and position on the lower side with the linkage always perpendicular to the passage.
The goal is spawned horizontally in a random position on the upper side. 

The disturbance in this scenario occurs in the form of one of the gaps narrowing enough to block the larger agent.
During this disturbance, the agents will need to perform a heterogeneous task assignment that directs them to their suitable gaps.

\subsubsection{Observation}

Each agent observes and communicates its velocity, relative position to each gap, and relative position to the goal center. The relative positions and velocities to the other agents are obtained through communication. The sizes of the agents or of the gaps are not part of the observations.

\subsubsection{Action}

Agents have a 2D continuous action force that determines their movement and they have the same action structure and mass.

\subsubsection{Reward}

The reward function is global and shared by the team. The agents are always rewarded for keeping the linkage parallel to the goal.
The navigation part of the reward is composed of two dense convex terms: before the passage, the agents are rewarded  to carry the linkage's center to the center of the passage; after the passage, the agents are rewarded for carrying it to the goal. Collisions are also penalized with a sparse reward.

\subsubsection{Termination}

The task is done when the agents reach the goal or after 300 steps. Each simulation step corresponds to 0.1 seconds.

\section{Additional tasks}
\begin{figure*}[t]
\centering
\includegraphics[width=\textwidth]{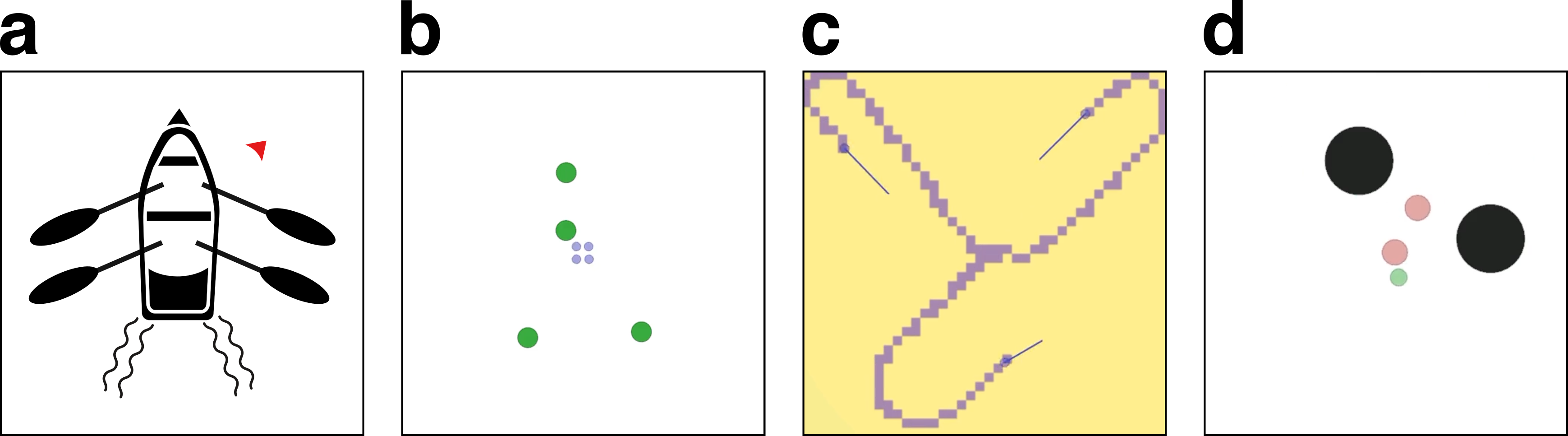}
\caption{\textbf{Additional tasks supporting the key role of diversity in collective learning.}
\textbf{a,} \textit{Differential steering}~\cite{bettini2023snd} requires four agents (two on each side) to collectively steer a boat. Results in this task show that learning can be bootstrapped when a prior on the diversity value is available analytically. 
\textbf{b,} \textit{Dispersion}~\cite{bettini2024controlling} requires agents (blue circles) to collectively consume food particles (green circles) in the environment. By controlling diversity we are able to stimulate agents to all tackle different food particles.
\textbf{c,} \textit{Sampling}~\cite{bettini2024controlling} requires agents (blue circles) to collectively sample a uniform map. Controlling diversity in this task enforces spatial separation of agents, leading to better performance.
\textbf{d,} \textit{Tag}~\cite{bettini2024controlling} requires two red chasing agents to tag the green escaping agents. We show that diversity in this task leads to the emergence of novel high-performing strategies.
}
\label{fig:additional_tasks}
\end{figure*}

In addition to the tasks presented in this paper, we summarize results form tasks presented in prior work, providing additional evidence of the key role of diversity in collective learning:
\begin{itemize}
    \item \textit{Differential steering}~\cite{bettini2023snd} requires four agents (two on each side) to collectively steer a boat. In this task the optimal diversity can be computed analytically and we are thus able to bootstrap the learning phase by directly enforcing that value for diversity.
    \item \textit{Dispersion}~\cite{bettini2024controlling} requires agents (blue circles) to collectively consume food particles (green circles) in the environment. By controlling diversity we are able to stimulate agents to all tackle different food particles.
    \item \textit{Sampling}~\cite{bettini2024controlling} requires agents (blue circles) to collectively sample a uniform map. Controlling diversity in this task enforces spatial separation of agents, leading to better performance.
    \item \textit{Tag}~\cite{bettini2024controlling} requires two red chasing agents to tag the green escaping agents. We show that diversity in this task leads to the emergence of novel high-performing strategies.
\end{itemize}

\subsection{\textit{Tag}: emergent adversarial strategies}
\label{sec:tag}

\begin{figure}[t]
\begin{center}
\centerline{\includegraphics[width=\linewidth]{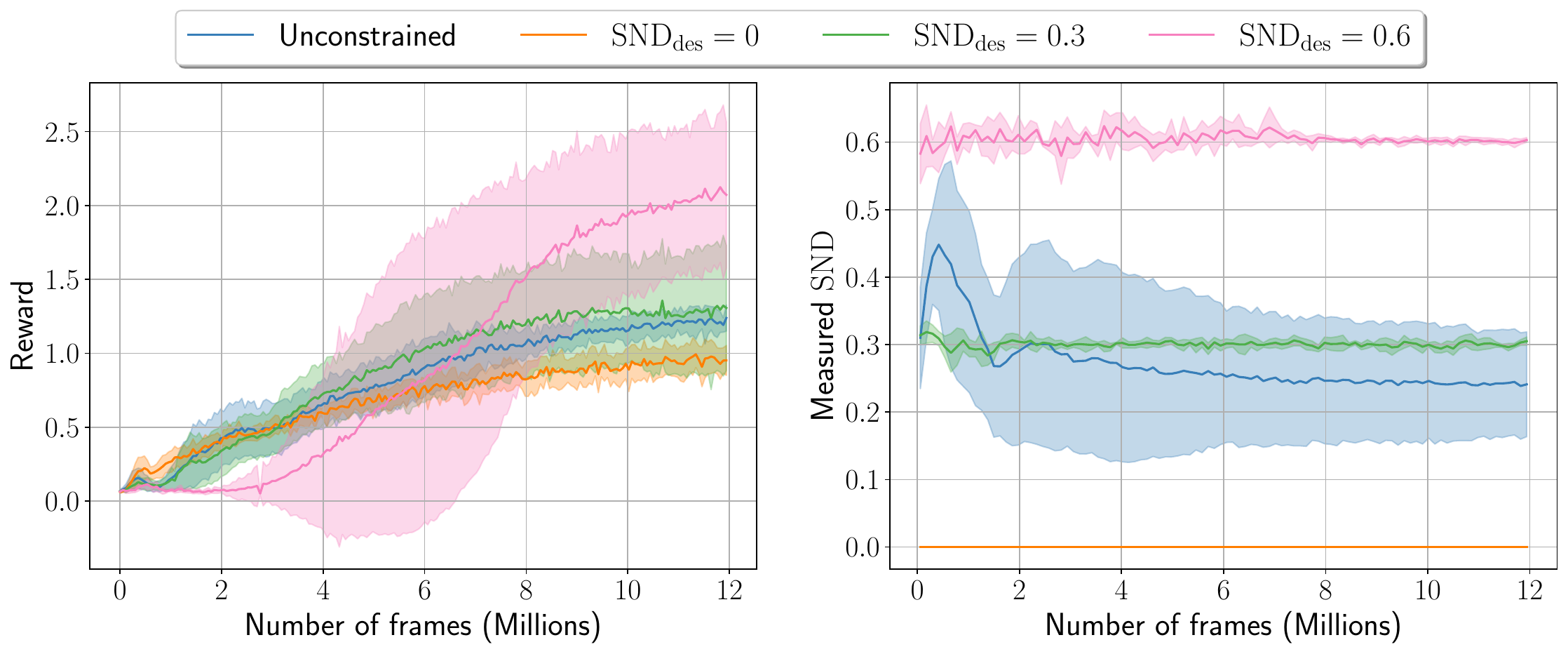}}
\caption{Results from training agents with different constraints on the \textit{Tag} task. \textbf{Left:} Mean instantaneous reward. \textbf{Right:} Measured diversity $\mathrm{SND}$. 
Curves report mean and standard deviation for the IPPO algorithm over 5 training seeds.}
\label{fig:tag_plots}
\end{center}
\end{figure}

In this section, we delve more in depth into the results of the \textit{Tag} experiments~\cite{bettini2024controlling} to illustrate how the concepts illustrated in this work can be leveraged in adversarial settings.

\textit{Tag}~\cite{lowe2017multi}(\autoref{fig:additional_tasks}d) is an adversarial task where a team of $n$ red agents is rewarded for tagging the green agent. All agents are spawned at random positions in a 2D space, alongside two randomly spawned obstacles (black circles). Agents can collide among each other and with obstacles.
Actions are 2D forces that determine their motion.
They observe their position and the relative positions to all other agents.
Red agents obtain a shared reward of 1 for each timestep where they are touching the green agent. The green agent gets a reward of -1 for the same condition. 

We train $n=2$ red agents using IPPO with different DiCo constraints, as well as unconstrained heterogeneous polices. The green agent is also trained with IPPO for 3 million frames, and frozen for the remainder of training.
In \autoref{fig:tag_plots}, we report the mean instantaneous training reward and the measured diversity of the red agents. From the plots, we observe that homogeneous agents ($\mathrm{SND}_\mathrm{des} = 0$) and unconstrained heterogeneous agents obtain similar rewards, with the heterogeneous model performing slightly better. We also train a DiCo constrained model at the diversity level obtained by the unconstrained one ($\mathrm{SND}_\mathrm{des} = 0.3$) and observe that this model achieves the same reward. By inspecting the rollouts of these policies, we note that they present a behavioral similarity: the red agents blindly chase the green one, all trying to minimize their distance to the target. In real-life ball sports, this is a well-known myopic and suboptimal strategy (\textit{e.g.}, an entire team chasing the ball in soccer). We can intuitively also see its suboptimality in this task: due to the shared nature of the tagging reward, the chasing agents could improve their spatial coverage by diversifying their strategies. To confirm our hypothesis, we perform experiments with a higher desired diversity ($\mathrm{SND}_\mathrm{des} = 0.6$). The results prove our hypothesis, with the constrained model able to almost double the obtained reward. The agents, constrained at this diversity, show the emergence of several fascinating new strategies that resemble strategies employed by human players in ball games (\textit{e.g.}, man-to-man marking, pinching maneuvers, spreading to cut off the evader).
The results demonstrate that, by constraining the policy search space to a specified diversity level, DiCo can be used to learn novel and diverse strategies that can overcome the suboptimality of unconstrained heterogeneous agents.

\begin{figure*}[!t]
    \centering
    \subfigure[$\mathrm{SND}_\mathrm{des}=0$.]{
        \includegraphics[width=\textwidth]{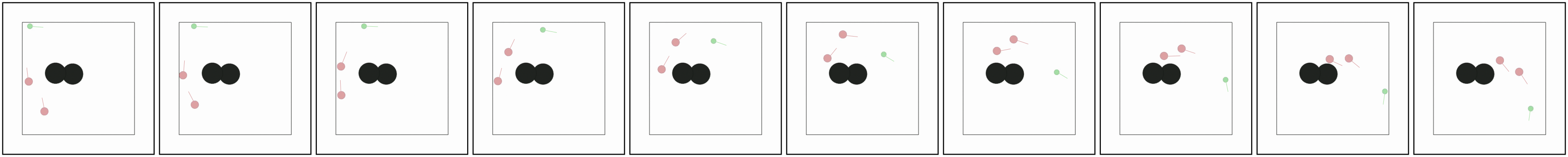}
        \label{fig:tag_0_traj}
    }
    \subfigure[Unconstrained.]{
        \includegraphics[width=\textwidth]{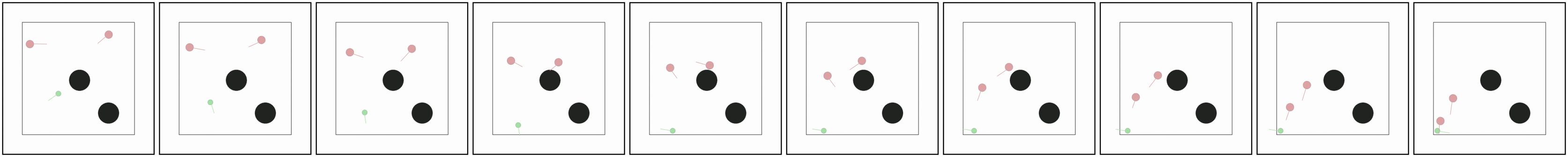}
        \label{fig:tag_-1_traj}
    }
    \subfigure[Ambush emergent strategy ({$\mathrm{SND}_\mathrm{des}=0.6$}).]{
        \includegraphics[width=\textwidth]{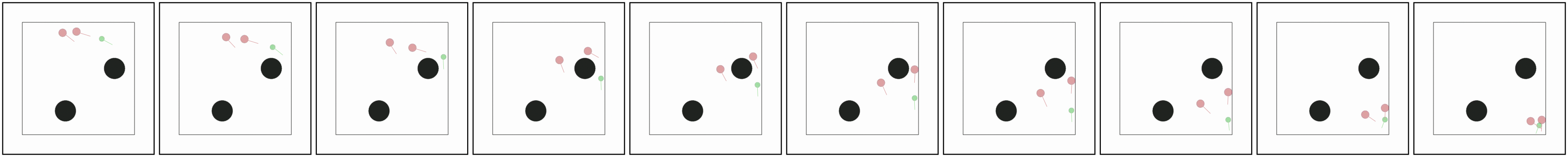}
        \label{fig:tag_ambush_traj}
    }
     \subfigure[Cornering emergent strategy ({$\mathrm{SND}_\mathrm{des}=0.6$}).]{
        \includegraphics[width=\textwidth]{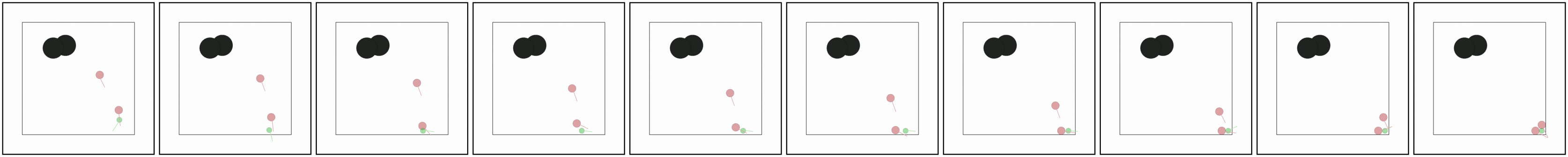}
        \label{fig:tag_cornering_traj}
    }
     \subfigure[Blocking emergent strategy ({$\mathrm{SND}_\mathrm{des}=0.6$}).]{
        \includegraphics[width=\textwidth]{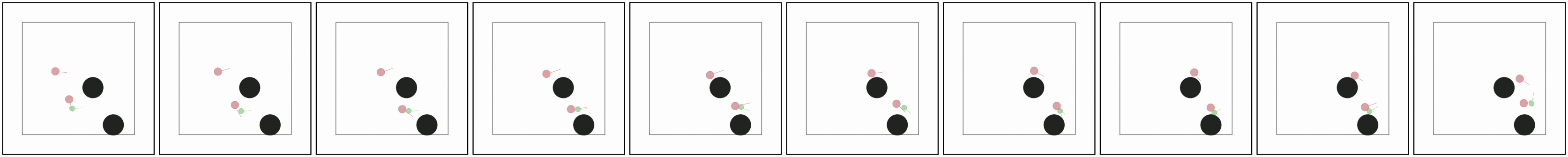}
        \label{fig:tag_blocking_traj}
    }
    \caption{\textit{Tag} trajectory renderings from the results in \autoref{fig:tag_plots}. Rollouts evolve from left to right. Videos of these trajectories are available on the DiCo paper \href{https://sites.google.com/view/dico-marl}{website}.}
    \label{fig:tag_traj}
\end{figure*}

We further analyze the renderings of the trajectories obtained in this experiment, reported in \autoref{fig:tag_traj}.

In \autoref{fig:tag_0_traj} and \autoref{fig:tag_-1_traj} we report the trajectories of homogeneous agents ($\mathrm{SND}_\mathrm{des}=0$) and unconstrained heterogeneous agents.
Both models present a similar strategy, consisting in both red agents navigating towards the green one, following the shortest path. This strategy is well known to be suboptimal in ball games like soccer, where inexperienced players tend to group on the ball, resulting in poor spatial coverage.

Continuing with this analogy, we take on a coaching role and constrain agents to a higher diversity ($\mathrm{SND}_\mathrm{des}=0.6$). We observe that this higher diversity constraint induces higher achieved rewards and emergent strategies that leverage agent complementarity. We highlight three examples of such strategies (\autoref{fig:tag_ambush_traj}, \autoref{fig:tag_cornering_traj}, \autoref{fig:tag_blocking_traj}).

In \autoref{fig:tag_ambush_traj}, we observe how the red agents are able to perform an ``ambush'' maneuver, splitting around the obstacle (from the \nth{3} frame). While the right agent performs a close chase, the left agent takes a longer trajectory, which, in the long term, enables it to cut off the green agent. We can see how diversity helped in this scenario: if both agents followed the evader into the passage, the green agent would have had a free escape route on the other side and the chase would have continued in a loop as in \autoref{fig:tag_0_traj}.

In \autoref{fig:tag_cornering_traj}, we observe how the red agents are able to coordinate in a pinching maneuver, cornering the evader in an inescapable state. While one agent lures the evader into a corner (frames 1 to 7), the other agent approaches, progressively closing the escape routes available to the green agent. Eventually, the green agent remains stuck in the corner without any possibility of movement.

Lastly, in \autoref{fig:tag_blocking_traj} we report a strategy where one red agent focuses on blocking the escape routes of the green agent, while the other red agent gathers rewards. In this strategy, more subtle to understand, the red agent in the top performs man-to-man marking (similar to what is done in sports) at a distance. As the green agent moves, this agent tracks its position, blocking access to the top half of the environment by being ready to intercept any movement in that direction.
In doing so, it effectively reduces the maneuvering space available to the green agent, making tagging easier for the other red agent.


\end{appendices}
\newpage

\bibliography{sn-bibliography}


\end{document}